\documentclass[letterpaper]{article} % DO NOT CHANGE THIS
\usepackage[preprint]{aaai2027}  % Public arXiv version
\usepackage[hyphens]{url}  % DO NOT CHANGE THIS
\usepackage{graphicx} % DO NOT CHANGE THIS
\usepackage{natbib}  % DO NOT CHANGE THIS AND DO NOT ADD ANY OPTIONS TO IT
\usepackage{caption} % DO NOT CHANGE THIS AND DO NOT ADD ANY OPTIONS TO IT
\usepackage{algorithm}
\usepackage{algorithmic}

\usepackage{newfloat}
\usepackage{listings}
\DeclareCaptionStyle{ruled}{labelfont=normalfont,labelsep=colon,strut=off} % DO NOT CHANGE THIS
\floatstyle{ruled}
\newfloat{listing}{tb}{lst}{}
\floatname{listing}{Listing}

\usepackage{booktabs}
\usepackage{amsmath,amssymb}
\nocopyright
\title{How Much Do $\ell_1$ Attacks Need Concentrated Perturbations?
}
\author{
    Melika Shirian\textsuperscript{\rm 1}\equalcontrib,
    Kianoosh Vadaei\textsuperscript{\rm 1}\equalcontrib,
    Arshia Hemmat\textsuperscript{\rm 2}\equalcontrib
}

\affiliations{
    \textsuperscript{\rm 1}Independent Researcher\\
    \textsuperscript{\rm 2}University of Oxford\\
    melika.shi@mehr.ui.ac.ir, k.vadaei@mehr.ui.ac.ir\\
    amirarshia.hemmat@kellogg.ox.ac.uk
}

\begin{document}

\maketitle

\begin{abstract}
A global $\ell_1$ budget limits how much an adversary may change an image,
not how that change is distributed. We ask how much of an $\ell_1$ attack's
observed success remains when large coordinate changes are restricted.
We add a per-coordinate cap to the budget, enforce the joint constraint
by exact in-loop projection, and sweep the cap for APGD-$\ell_1$, SLIDE,
and FMN-$\ell_1$ without altering their objectives. Across 18 standard-model
settings, caps of at most $32/255$ retain a median $97\%$ of attack success
relative to each method's native uncapped baseline, while median peak
magnitude falls by $80\%$ and median effective support nearly triples.
At matched realized $\ell_1$, capped perturbations remain structurally
distinct; separate ImageNet comparisons also show lower LPIPS and DISTS
at selected caps. Direct in-loop constrained optimization preserves up
to $88$ percentage points more attack success than post-hoc clipping,
highlighting the importance of constraint enforcement. Suitable caps
nevertheless vary across attacks and models, and adversarially trained
classifiers exhibit more heterogeneous concentration sensitivity.
These results establish perturbation concentration as a complementary,
model- and attack-dependent axis of $\ell_1$ attack evaluation.
\end{abstract}

\begin{figure}[!t]
    \centering
    \includegraphics[width=\columnwidth]{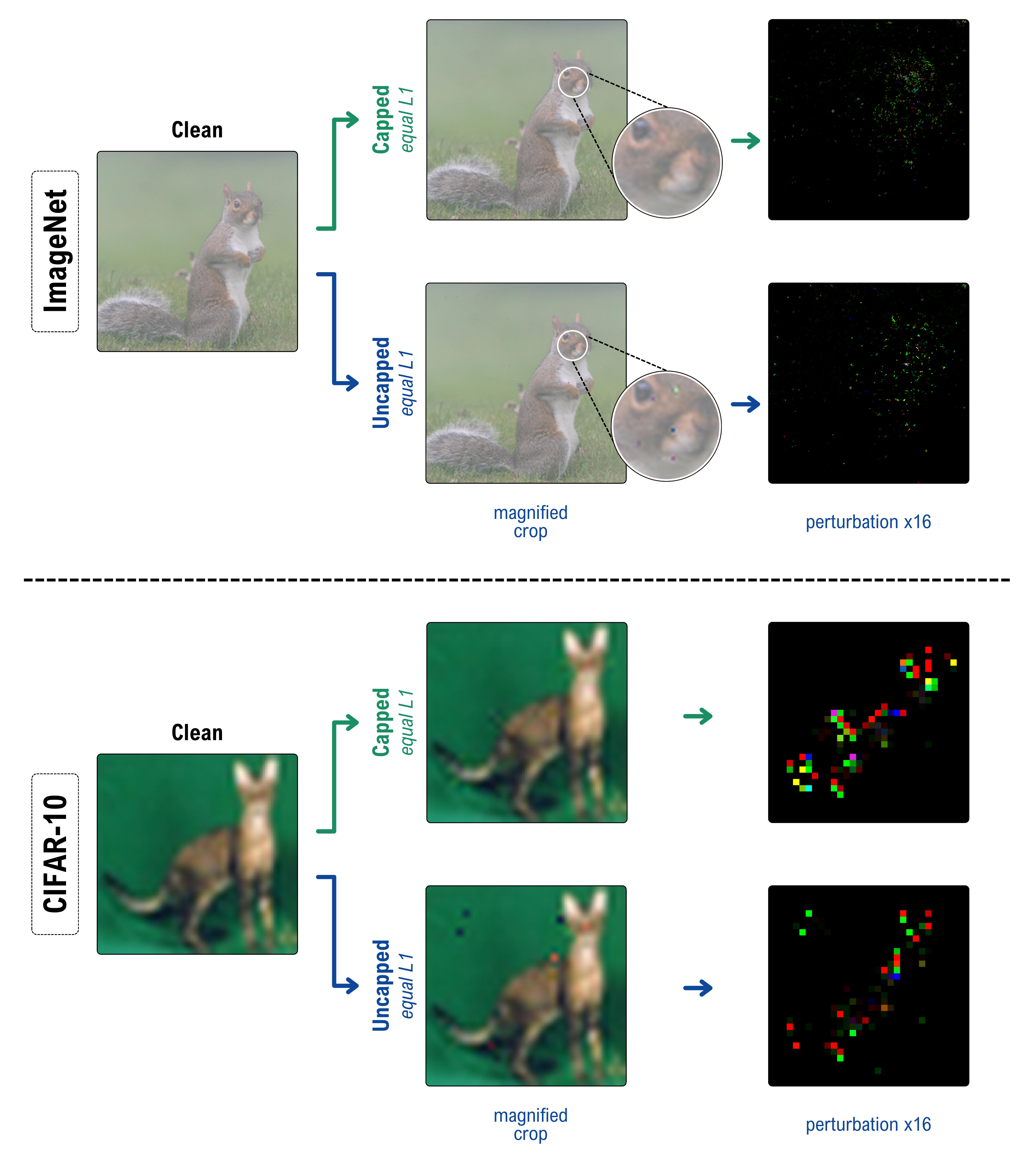}
    \caption{\textbf{Equal-$\ell_1$ perturbations can have substantially
    different local structures.} APGD-$\ell_1$ examples on
    ImageNet/ResNet-50 (top) and CIFAR-10/ResNet-18 (bottom) compare capped and
    conventional uncapped ($c=1$) perturbations matched by realized $\ell_1$
    magnitude. Both are successful, but the capped perturbations have lower
    peak changes and broader support. Magnified crops highlight these
    differences. Perturbations are amplified by $16\times$ for visualization
    and do not represent their original-scale appearance.}
    \label{fig:method_overview}
\end{figure}

\section{Introduction}
\label{sec:intro}

% \label{sec:introduction}
% ============================================================

Adversarial robustness is typically evaluated by constraining perturbations to
a norm-bounded set and measuring attack success
\cite{szegedy2014intriguing,madry2018towards}. Although this convention enables
reproducible comparisons across attacks and models
\cite{croce2020reliable,croce2021robustbench}, a global $\ell_1$ budget
constrains total perturbation magnitude without specifying how that magnitude
is distributed. The same total mass may be concentrated on a few coordinates
or spread across many, producing perturbations that are equivalent in
$\ell_1$ norm but differ substantially in peak magnitude, effective support,
spatial structure, and perceptual distortion.

The tendency of $\ell_1$ attacks to produce sparse, concentrated perturbations
is well documented \cite{tramer2019adversarial,croce2021mind}. What remains
unclear is how much of their observed effectiveness survives restrictions on
large coordinate-wise changes. We therefore ask:
\emph{what happens when large local modifications are restricted while the
global $\ell_1$ budget remains available?} Can an attack redistribute its
perturbation and remain effective, and how does this sensitivity vary across
attacks and models?

We investigate four aspects of this question: \textbf{(Q1)} how attack success
changes as coordinate-wise modifications are progressively restricted;
\textbf{(Q2)} whether, at comparable realized $\ell_1$ magnitude, the
restriction changes peak magnitude, effective support, and top-mass
concentration rather than merely reducing budget usage; \textbf{(Q3)} whether
capped operating points exhibit lower measured perceptual distances while
retaining attack success; and \textbf{(Q4)} how the findings vary across
attacks, optimization settings, datasets, architectures, and model-robustness
regimes.

To study this, we augment the conventional global $\ell_1$ budget with a
per-coordinate magnitude cap $c$. The uncapped condition $c=1$ recovers the
standard image-domain $\ell_1$ setting, while smaller caps define nested
restrictions of the same threat set. Unlike conventional $\ell_\infty$-only
evaluation, which bounds coordinate-wise changes without independently
constraining total perturbation mass, our protocol holds the global $\ell_1$
budget fixed while varying the permitted peak magnitude. This characterizes
concentration sensitivity within a shared-budget threat model rather than
merely comparing different single-norm constraints. Sweeping $c$ produces
a perturbation--concentration frontier, summarized by the tightest evaluated
cap retaining at least $95\%$ of the native uncapped attack success.
Because the feasible sets are nested, empirical ASR increases under tighter
caps reflect finite-compute optimization rather than stronger theoretical
threat sets. Formal definitions are given in Sec.~\ref{sec:approach}.

We enforce the cap during optimization using exact projection onto the
intersection of the $\ell_1$ ball, coordinate cap, and valid image box.
This established projection is incorporated into APGD-$\ell_1$, SLIDE,
and FMN-$\ell_1$ without changing their attack objectives. Native uncapped
implementations serve as reference conditions, while capped conditions
use the joint projector. In particular, SLIDE's projection operator differs
between these conditions, so its native-relative retention does not isolate
the causal effect of the cap alone. Our contribution is therefore an
evaluation protocol rather than a new attack family; post-hoc clipping
serves only as a control because it prevents perturbation redistribution.

Across the primary budgets of the standard models, all 18
attack--model--dataset settings retain at least $95\%$ of their native
uncapped ASR at an evaluated cap no larger than $32/255$, with selected
caps ranging from $2/255$ to $32/255$. Matched-realized-$\ell_1$ analyses
further show that capped perturbations can exhibit lower peak magnitude,
broader effective support, and less mass concentrated in their largest
coordinates without merely using less total perturbation. Selected
ImageNet/ResNet-50 settings also achieve lower measured LPIPS and DISTS
\cite{zhang2018unreasonable,ding2020image}. The frontiers nevertheless
vary substantially across attacks and models, and adversarially trained
models show more heterogeneous concentration sensitivity, ruling out a
universal cap.

Our controls further characterize this finding. Direct in-loop constrained
optimization achieves substantially higher ASR than post-hoc clipping under
the evaluated conditions, and stronger optimization profiles preserve
high-retention behavior in the tested configurations. Repeated-search
audits additionally illustrate the role of finite-compute optimization
in observed non-monotonic behavior. Together, these observations
characterize concentration sensitivity as an empirical property that
varies with attack and model configurations.

Our contributions are:

\begin{itemize}
    \item We formulate concentration-aware $\ell_1$ evaluation using nested
    $\ell_1$--$\ell_\infty$ threat sets, empirical concentration frontiers,
    and retention-based operating points relative to native attack baselines.

    \item We integrate established exact joint projection into three
    $\ell_1$ attack families and provide controls examining perturbation
    redistribution, budget under-use, post-hoc clipping, and sensitivity
    to optimization strength.

    \item Across CIFAR-10, CIFAR-100, and ImageNet, standard and adversarially
    trained models, and convolutional and transformer architectures, we
    characterize when lower-peak, more distributed perturbations preserve
    empirical attack success and where this behavior becomes model- and
    attack-dependent.
\end{itemize}

Together, these results show that a global $\ell_1$ budget specifies how much
an adversary may change, but not how successful attacks allocate that change.
Systematically varying the permitted concentration reveals differences in
empirical attack effectiveness and perturbation structure that a single
uncapped evaluation does not capture.

\section{Related work}
\label{sec:related-work}

\paragraph{Sparse and $\ell_1$ adversarial attacks.}
Since adversarial examples were first characterized
\cite{szegedy2014intriguing,goodfellow2015explaining}, most work has focused on
$\ell_\infty$ and $\ell_2$ threat models
\cite{madry2018towards,carlini2017towards}. Sparse attacks instead restrict the
number of modified coordinates, from saliency-based
\cite{papernot2016limitations} and single-pixel attacks
\cite{su2019onepixel} to geometric \cite{modas2019sparsefool}, structured
\cite{fan2020sparse}, and query-efficient black-box methods
\cite{croce2022sparsers}. The $\ell_1$ ball allows sparsity without fixing the
support and underlies elastic-net \cite{chen2018ead}, gradient-sparsified
\cite{tramer2019adversarial}, and adaptive projected-gradient attacks
\cite{croce2021mind,croce2020reliable}. Related methods instead minimize
adversarial norm \cite{rony2019decoupling,croce2020minimally,pintor2021fast,rony2021augmented}.
Efficient $\ell_1$ projection \cite{duchi2008efficient,condat2016fast} is
therefore central, and exact projection onto the intersection of the
$\ell_1$ ball and valid image box has been established
\cite{croce2021mind}. In these settings, the global budget is specified,
but how perturbation mass is distributed across coordinates is largely
left to the optimizer.

\paragraph{Perceptual constraints and perturbation allocation.}
$\ell_p$ proximity is not equivalent to perceptual similarity
\cite{sharif2018suitability}, and its role as a security model has been
questioned \cite{gilmer2018motivating}. Alternatives use learned perceptual
distances \cite{laidlaw2021perceptual}, transport costs
\cite{wong2019wasserstein}, unions of norms
\cite{tramer2019adversarial,maini2020adversarial}, unforeseen distortions
\cite{kaufmann2023testing}, or spatial and semantic transformations
\cite{xiao2018spatially,engstrom2019exploring,hosseini2018semantic}.
Closer to our setting, additive sparse attacks promote less visible
perturbations through local-variation constraints \cite{croce2019sparse},
learned distortion maps \cite{dong2020greedyfool}, factorization
\cite{fan2020sparse}, or homotopy optimization \cite{zhu2021sparse}.
Other approaches explicitly control perturbation sparsity and allocation
during attack construction. SAIF \cite{imtiaz2025saif} jointly optimizes
an $\ell_1$-bounded soft mask and an $\ell_\infty$-bounded perturbation using Frank--Wolfe, whereas a complementary importance guided approach \cite{shirian2026dontwaste} uses a fixed clean-gradient prior to redistribute perturbation mass toward model-sensitive coordinates.
These approaches
demonstrate the relevance of perturbation allocation to attack design,
although their formulations and optimization mechanisms differ.

Our focus is complementary: rather than optimizing a new allocation
mechanism, we examine how the empirical effectiveness of established
$\ell_1$ attacks changes as the permitted coordinate magnitude is
progressively restricted. We preserve their objectives and update rules,
enforce the joint $\ell_1$--cap--box constraint in-loop, and sweep the cap
at fixed global budget. Unlike an $\ell_\infty$-only evaluation, this
retains an independent constraint on total perturbation mass. The
resulting nested-set frontier characterizes concentration sensitivity
relative to native uncapped attack baselines. Here, concentration denotes
within-perturbation mass allocation, not concentration of measure
\cite{mahloujifar2019curse}; SSIM \cite{wang2004image}, LPIPS
\cite{zhang2018unreasonable}, and DISTS \cite{ding2020image} are related
perceptual measures, with LPIPS and DISTS used only as supporting
measurements in our experiments.

\paragraph{Evaluating adversarial robustness.}
Robustness estimates depend critically on evaluation quality: weak attacks
can overestimate robustness \cite{uesato2018adversarial}, while gradient
masking can fail under adaptive analysis
\cite{athalye2018obfuscated,tramer2020adaptive}. This has motivated evaluation
guidelines \cite{carlini2019evaluating}, parameter-free ensembles
\cite{croce2020reliable}, standardized benchmarks \cite{croce2021robustbench},
minimum-distortion estimators \cite{brendel2019accurate}, shared toolkits
\cite{rauber2017foolbox}, and optimization-failure diagnostics
\cite{pintor2022indicators}, including for adversarially trained models
\cite{madry2018towards,bartoldson2024adversarial}. We study a complementary
source of evaluation variation: feasible-set geometry and enforcement.
Our cap sweep measures empirical concentration sensitivity, while the
large gap between direct in-loop constrained optimization and post-hoc
clipping---up to $88$ percentage points---highlights the importance of
optimizing within the intended feasible set rather than modifying
completed perturbations.

\begin{table}[t]
    \centering
    \caption{
    Main perturbation--concentration results for standard models at the
    primary budgets ($B=4$ for CIFAR-10/100 and $B=20$ for ImageNet).
    $c^\star$ is the tightest evaluated cap retaining at least $95\%$ of the
    native uncapped ASR. $\Delta_\infty$ denotes the reduction in median
    $\ell_\infty$, and $\Delta S_{\mathrm{eff}}$ the increase in median
    effective support.
    }
    \label{tab:main_concentration_standard}
    \scriptsize
    \setlength{\tabcolsep}{1.6pt}
    \resizebox{\columnwidth}{!}{%
    \begin{tabular}{lllrrrrrr}
        \toprule
        Data & Model & Attack &
        $255c^\star$ &
        ASR$_1$ &
        ASR$_\star$ &
        Ret. &
        $\Delta_\infty$ &
        $\Delta S_{\rm eff}$ \\
        & & & & \multicolumn{5}{c}{(\%)} \\
        \midrule

        C10 & ResNet-18 & APGD-L1
        & 16 & 96.04 & 92.65 & 96.47 & 79.80 & 145.94 \\
        C10 & ResNet-18 & SLIDE
        & 32 & 90.80 & 88.60 & 97.58 & 56.31 & 46.17 \\
        C10 & ResNet-18 & FMN-L1
        & 16 & 95.98 & 92.29 & 96.16 & 72.77 & 186.79 \\

        \addlinespace

        C10 & WRN-28-10 & APGD-L1
        & 8 & 96.91 & 93.02 & 95.98 & 88.28 & 268.71 \\
        C10 & WRN-28-10 & SLIDE
        & 32 & 89.31 & 88.09 & 98.63 & 48.40 & 31.17 \\
        C10 & WRN-28-10 & FMN-L1
        & 16 & 96.40 & 94.97 & 98.51 & 66.85 & 137.30 \\

        \addlinespace

        C100 & ResNet-18 & APGD-L1
        & 8 & 97.48 & 93.88 & 96.30 & 89.44 & 324.58 \\
        C100 & ResNet-18 & SLIDE
        & 32 & 93.34 & 91.75 & 98.31 & 48.40 & 31.15 \\
        C100 & ResNet-18 & FMN-L1
        & 8 & 96.15 & 92.41 & 96.11 & 80.89 & 339.96 \\

        \addlinespace

        C100 & WRN-28-10 & APGD-L1
        & 2 & 97.38 & 98.26 & 100.90 & 97.37 & 1032.49 \\
        C100 & WRN-28-10 & SLIDE
        & 8 & 91.40 & 87.29 & 95.50 & 73.43 & 61.08 \\
        C100 & WRN-28-10 & FMN-L1
        & 2 & 92.22 & 89.87 & 97.45 & 89.82 & 779.65 \\

        \addlinespace

        IN & ResNet-50 & APGD-L1
        & 4 & 89.03 & 84.99 & 95.45 & 96.69 & 742.95 \\
        IN & ResNet-50 & SLIDE
        & 8 & 86.29 & 83.42 & 96.67 & 79.66 & 96.19 \\
        IN & ResNet-50 & FMN-L1
        & 8 & 89.69 & 88.77 & 98.98 & 84.19 & 239.74 \\

        \addlinespace

        IN & ViT-B/16 & APGD-L1
        & 4 & 63.50 & 67.25 & 105.91 & 96.47 & 649.35 \\
        IN & ViT-B/16 & SLIDE
        & 8 & 71.12 & 68.25 & 95.96 & 73.75 & 59.51 \\
        IN & ViT-B/16 & FMN-L1
        & 16 & 76.00 & 74.00 & 97.37 & 78.07 & 166.34 \\

        \bottomrule
    \end{tabular}%
    }

\end{table}

\section{Concentration-aware evaluation}
\label{sec:approach}

% \section{Concentration-Aware Evaluation}

We ask how much of a conventional $\ell_1$ attack's observed effectiveness
survives when large changes to individual image coordinates are
progressively restricted. The global $\ell_1$ budget is held fixed, the
permitted coordinate magnitude is swept, and standard attack success is
measured at each condition. The protocol introduces no new attack objective
and no new definition of success; it augments conventional $\ell_1$
evaluation with an explicit concentration axis that a single uncapped
ASR measurement cannot recover.

\subsection{Concentration-controlled $\ell_1$ evaluation}
\label{sec:controlled_evaluation}

Let $x\in[0,1]^D$ be an image and $\delta\in\mathbb{R}^D$ an additive
perturbation. For a global $\ell_1$ budget $B$ and a per-coordinate magnitude
cap $c$, the admissible set is
\begin{equation}
\label{eq:approach_threat_set}
\mathcal{S}(B,c;x)=
\left\{
\begin{aligned}
\delta\in\mathbb{R}^{D}:\quad
&\|\delta\|_1\leq B,\\
&\|\delta\|_\infty\leq c,\\
&x+\delta\in[0,1]^D
\end{aligned}
\right\}.
\end{equation}
Here $B$ limits the total absolute perturbation magnitude while $c$ limits
the magnitude assigned to any individual coordinate, including its channel
index; all constraints and measurements are defined in raw $[0,1]$ image
space.

Setting $c=1$ recovers the conventional image-domain $\ell_1$ threat set,
since $x+\delta\in[0,1]^D$ already prevents any coordinate from changing by
more than one. Smaller caps define controlled restrictions of that set: for
$c_1\leq c_2$,
\begin{equation}
\label{eq:approach_nested_sets}
\mathcal{S}(B,c_1;x)
\subseteq
\mathcal{S}(B,c_2;x)
\subseteq
\mathcal{S}(B,1;x).
\end{equation}
A capped condition is therefore deliberately more restrictive, and we never
treat capped and uncapped conditions as equivalent adversarial capabilities.
The $c=1$ condition anchors the conventional evaluation; the remaining
conditions measure how much of its empirically observed effectiveness
survives once access to large coordinate-wise changes is withdrawn.
Holding $B$ fixed preserves the nominal global perturbation budget while
varying the permitted coordinate magnitude.

\subsection{Retained effectiveness}
\label{sec:effectiveness_retention}

Let $\mathcal{A}$ be an attack, $f$ a classifier, and $\mathcal{C}$ an
ordered cap grid containing $1$. For each $c\in\mathcal{C}$ we run
$\mathcal{A}$ under $\mathcal{S}(B,c;x)$ and compute the conventional attack
success rate on the same clean-correct population, written
$\operatorname{ASR}_{\mathcal{A},f}(B,c)$; the arguments specify the feasible
region, not a different success criterion. The
perturbation--concentration frontier, its retention relative to the
native uncapped execution, and the resulting operating point are
\begin{equation}
\label{eq:approach_frontier}
F_{\mathcal{A},f,B}(c)=\operatorname{ASR}_{\mathcal{A},f}(B,c),
\qquad c\in\mathcal{C},
\end{equation}
\begin{equation}
\label{eq:approach_retention}
R_{\mathcal{A},f,B}(c)=
\frac{F_{\mathcal{A},f,B}(c)}{F_{\mathcal{A},f,B}(1)},
\quad
c^{\star}_{\rho,\mathcal{A},f}(B)=
\min\{c\in\mathcal{C}: R\geq\rho\}.
\end{equation}
A flat region of the frontier indicates that large coordinate changes can
be restricted at little empirical cost relative to the reference execution;
a sharp decline indicates greater sensitivity under the evaluated
optimization configuration. Our primary summary uses $\rho=0.95$, so
$c^\star_{0.95}$ is the tightest \emph{evaluated} cap preserving at least
$95\%$ of the directly measured native uncapped ASR for the same attack,
model, budget, and population. This reference-relative statistic
characterizes the specified attack implementations rather than
best-achievable adversarial success across optimization configurations.
The full frontier remains the object of analysis; $c^\star$ is a compact
summary, selected independently for every attack--model--budget
configuration and never transferred between settings. We additionally
examine $\rho\in\{0.90,0.99\}$.

Because the sets in Eq.~\eqref{eq:approach_nested_sets} are nested,
tightening the cap cannot increase the exact worst-case adversarial risk.
Finite-compute attacks may nevertheless produce non-monotonic empirical ASR,
because changing the constraint changes the geometry the optimizer
encounters. We therefore report measured frontiers without monotonic
smoothing and interpret any local increase under a tighter cap as
optimization behavior rather than evidence that the smaller threat set
is stronger.

\subsection{Realizing the capped conditions}
\label{sec:constraint_realization}

The cap must be enforced during optimization. Clipping a completed uncapped
perturbation discards mass without letting the optimizer reallocate it, so
it cannot answer whether the same attack could have found a successful,
less concentrated perturbation inside the joint feasible set. We therefore
enforce the constraints by Euclidean projection: for a proposed
perturbation $v$,
\begin{equation}
\label{eq:approach_projection}
\begin{aligned}
\Pi_{B,c,x}(v)
&=
\arg\min_z
\tfrac{1}{2}\|z-v\|_2^2
\\
\mathrm{s.t.}\quad
&\|z\|_1\leq B,
\qquad
\ell_i\leq z_i\leq u_i,
\end{aligned}
\end{equation}
with $\ell_i=\max(-c,-x_i)$ and $u_i=\min(c,1-x_i)$ jointly enforcing the
coordinate cap and the valid image domain. The solution is a
box-constrained soft threshold,
\begin{equation}
\label{eq:approach_projection_solution}
z_i(\lambda)
=
\operatorname{clip}
\!\left(
\operatorname{sign}(v_i)
\left[|v_i|-\lambda\right]_+,
\ell_i,
u_i
\right),
\end{equation}
where $[a]_+=\max(a,0)$. If the box projection at $\lambda=0$ already
satisfies the global budget it is returned unchanged; otherwise $\lambda$
is obtained by one-dimensional bisection. The derivation and its numerical
validation are given in App.~\ref{app:projection_derivation}.

For fixed-budget iterative attacks the projection is applied to the initial
perturbation and after every update; for adaptive minimum-norm attacks it
replaces the feasibility projection at the attack's current radius. In both
cases the objective, gradient computation, candidate selection, and
remaining update logic are left unchanged.

The $c=1$ condition uses each attack's native uncapped $\ell_1$
implementation, including SLIDE's released projector, whereas $c<1$
conditions use the exact joint projector. Accordingly, retention in
Eq.~\eqref{eq:approach_retention} is measured relative to native
implementations. For SLIDE, both the cap and projection operator change,
so its frontier does not isolate the causal effect of restricting
coordinate magnitudes. We therefore interpret these results descriptively
as native-relative retention. Post-hoc clipping remains a control rather
than a realization of the capped optimization procedure.

\subsection{Matched-realized-$\ell_1$ comparison}
\label{sec:matched_realized_l1}

Equal nominal budgets do not guarantee equal realized magnitude: an attack
may use substantially less than $B$, and tightening $c$ may change how
much. A capped perturbation could then appear less concentrated simply
because it spends less. We control for this per example. For a successful
capped perturbation $\delta_{\mathcal{A},n}^{B,c}$ we record its realized
magnitude and rerun the uncapped attack on the same sample with that value
as its budget,
\begin{equation}
\label{eq:approach_matched_l1}
L_n^{(c)}
=
\bigl\|\delta_{\mathcal{A},n}^{B,c}\bigr\|_1,
\qquad
\widetilde{\delta}_{\mathcal{A},n}^{(c)}
=
\delta_{\mathcal{A},n}^{L_n^{(c)},1}.
\end{equation}
The two perturbations share an input and approximately the same realized
$\ell_1$ magnitude but are generated under their respective capped and
native uncapped optimization procedures. The comparison is conditioned
on capped-source success, so the matched execution's success rate is a
diagnostic on that population rather than a dataset-level ASR, and
structural quantities are compared per example. If the structural
difference largely disappears after matching, it can be explained by
reduced total perturbation use; if it persists, it provides evidence
of different mass allocation beyond differences in realized $\ell_1$
magnitude alone.

\subsection{Describing perturbation allocation}
\label{sec:allocation_measures}

Retention says nothing about how successful perturbations change. We
therefore record realized $\ell_1$ magnitude alongside peak magnitude
$\|\delta\|_\infty$, the share of mass carried by the largest coordinates,
and effective support. Letting $\operatorname{Top}_q(\delta)$ index the
largest $\lceil qD\rceil$ values of $|\delta|$,
\begin{equation}
\label{eq:approach_topmass}
M_q(\delta)
=
\frac{\sum_{i\in\operatorname{Top}_q(\delta)}|\delta_i|}
{\|\delta\|_1+\varepsilon},
\qquad
S_{\mathrm{eff}}(\delta)
=
\frac{\|\delta\|_1^2}{\|\delta\|_2^2+\varepsilon},
\end{equation}
where $\varepsilon$ is a small numerical stabilizer and we report
$q\in\{0.01,0.05\}$. For a fixed amount of $\ell_1$ mass, the participation
ratio $S_{\mathrm{eff}}$ increases as that mass is distributed across more
coordinates and decreases as it concentrates on fewer.

Lower $\|\delta\|_\infty$, lower top-mass fractions, and larger
$S_{\mathrm{eff}}$ are complementary evidence of broader allocation. We
report them separately rather than collapsing them into a single
concentration score, because they capture different structural properties
and need not change by the same amount. At the selected operating point
we summarize the change from the uncapped condition by the percentage
reduction in median $\|\delta\|_\infty$ and the percentage increase in median
$S_{\mathrm{eff}}$, defined in App.~\ref{app:structural_defs}; these are the
structural quantities reported alongside ASR retention in the main table.

\subsection{Perceptual measurements}
\label{sec:perceptual_measurement}

Broader allocation does not by itself imply lower perceptual distortion, so
we treat perceptual similarity as a separate measurement rather than a
consequence of reduced concentration. On ImageNet we compute LPIPS and DISTS
between clean and adversarial images at the conventional uncapped condition,
the independently selected $c^\star$, and a tighter neighboring cap. Because
the population of successful attacks can change across conditions, we also
perform paired common-success comparisons, ensuring the result is not driven
by differing subsets of successfully attacked examples. Both are learned
image-distance measures: a decrease supports a claim of lower measured
perceptual distortion, not of human imperceptibility, visibility thresholds,
or perceptual preference.

The protocol is ordered so that $c^\star$ is determined only from ASR
retention on the predefined frontier, with every subsequent analysis used
to interpret and stress-test that selection rather than to choose it; the
complete ordering is given in App.~\ref{app:protocol}. It thereby separates
three questions that a single conventional $\ell_1$ ASR measurement cannot
resolve: how much measured attack effectiveness remains when large
coordinate changes are restricted, how successful perturbations redistribute
their mass, and whether that structural difference persists at matched
realized $\ell_1$. The datasets, models, attacks, budgets, cap grid, and
populations used to instantiate the protocol are specified in
Sec.~\ref{sec:setup}.

\begin{table}[t]
    \centering
    \caption{
    Perturbation--concentration results for adversarially trained models
    at their reference budgets ($B=4$ for CIFAR-10 and $B=20$ for ImageNet).
    $c^\star$ is the tightest evaluated cap retaining at least $95\%$ of
    the native uncapped ASR. $\Delta_\infty$ denotes the percentage reduction
    in median $\ell_\infty$, and $\Delta S_{\mathrm{eff}}$ the percentage
    increase in median effective support. C10 and IN denote CIFAR-10
    and ImageNet, respectively.
    }
    \label{tab:main_concentration_robust}
    \scriptsize
    \setlength{\tabcolsep}{1.6pt}
    \resizebox{\columnwidth}{!}{%
    \begin{tabular}{lllrrrrrr}
        \toprule
        Data & Model & Attack &
        $255c^\star$ &
        ASR$_1$ &
        ASR$_\star$ &
        Ret. &
        $\Delta_\infty$ &
        $\Delta S_{\rm eff}$ \\
        & & & & \multicolumn{5}{c}{(\%)} \\
        \midrule

        C10 & CH PreActRN-18 ($\ell_1$) & APGD-L1
        & 32 & 8.65 & 8.35 & 96.55 & 50.68 & 52.74 \\
        C10 & CH PreActRN-18 ($\ell_1$) & SLIDE
        & 64 & 7.80 & 7.67 & 98.38 & 1.10 & 4.29 \\

        \addlinespace

        C10 & MSD PreActRN-18 & APGD-L1
        & 64 & 12.14 & 11.96 & 98.59 & 9.25 & 4.47 \\
        C10 & MSD PreActRN-18 & SLIDE
        & 64 & 12.82 & 12.78 & 99.71 & 1.04 & 3.00 \\

        \addlinespace

        C10 & XCiT-S12 ($\ell_\infty$) & APGD-L1
        & 255 & 28.50 & 28.50 & 100.00 & 0.00 & 0.00 \\
        C10 & XCiT-S12 ($\ell_\infty$) & SLIDE
        & 64 & 21.00 & 20.46 & 97.46 & 15.76 & 6.67 \\

        \addlinespace

        IN & XCiT-S12 ($\ell_\infty$) & APGD-L1
        & 255 & 20.79 & 20.79 & 100.00 & 0.00 & 0.00 \\
        IN & XCiT-S12 ($\ell_\infty$) & SLIDE
        & 4 & 6.02 & 6.02 & 100.00 & 0.00 & 0.00 \\

        \bottomrule
    \end{tabular}%
    }
\end{table}

\section{Experimental setup}
\label{sec:setup}

\paragraph{Datasets, models, and evaluation populations.}
Standard-model frontiers use the complete $10{,}000$-image CIFAR-10 and
CIFAR-100 test sets \cite{krizhevsky2009learning}. For ResNet-18 and
WideResNet-28-10 \cite{he2016resnet,zagoruyko2016wide}, the clean-correct
populations are $(9{,}497,9{,}478)$ on CIFAR-10 and
$(7{,}592,8{,}212)$ on CIFAR-100; WideResNet checkpoints are obtained from
RobustBench \cite{croce2021robustbench}. On ImageNet
\cite{russakovsky2015imagenet}, a deterministic class-balanced manifest with
one validation image per class is shared by torchvision ResNet-50 and
ViT-B/16 \cite{he2016resnet,dosovitskiy2021vit}, yielding $766$ and $800$
clean-correct examples. We additionally evaluate the Croce--Hein
$\ell_1$-robust PreActResNet-18 \cite{croce2021mind}, MSD multi-norm
PreActResNet-18 \cite{maini2020adversarial}, and $\ell_\infty$-robust
XCiT-S12 \cite{debenedetti2023light} on CIFAR-10, plus robust XCiT-S12 on
ImageNet, with clean-correct populations $8{,}706$, $8{,}174$, $9{,}006$, and
$731$, respectively. Attacks operate in the $[0,1]$ image domain with
normalization inside the classifier. ASR uses a common clean-correct
population for each dataset--model pair; success requires a label change and
satisfaction of all $\ell_1$, cap, and image constraints. Manifest
construction and checkpoint provenance are detailed in
Apps.~\ref{app:data} and~\ref{app:models}.

\paragraph{Frontier attacks.}
Standard frontiers use untargeted APGD-L1 \cite{croce2021mind}, SLIDE
\cite{tramer2019adversarial}, and FMN-L1 \cite{pintor2021fast}, instantiated
from official AutoAttack, the released SLIDE design, and authors-maintained
Foolbox, respectively. Each attack uses its native path at $c=1$; for $c<1$,
only projection is replaced by the exact joint operator from
Sec.~\ref{sec:constraint_realization}. SLIDE retains its margin objective,
boundary masking, Laplace initialization, $99$th-percentile sparsification,
and step size $0.255B$; its reported $c=1$ rows are corrected independent
native-projection reruns. FMN-L1 is run once per cap and evaluated along its
minimum-norm trajectory. All attacks use $100$ steps; APGD-L1 and SLIDE use
one restart and FMN-L1 one trajectory. Corresponding caps share deterministic
randomness. Robust frontiers use APGD-L1 and SLIDE only. Complete attack
instantiations are given in App.~\ref{app:attacks}.

\paragraph{Budgets, caps, and operating points.}
Standard budgets are $B\in\{2,4\}$ on CIFAR-10/100 and
$B\in\{10,20\}$ on ImageNet. Robust budgets are $\{4,12\}$ for Croce--Hein
and MSD, $\{2,4\}$ for CIFAR-10 XCiT-S12, and $\{20,255\}$ for ImageNet
XCiT-S12; non-reference robust budgets are interpreted only within model
(App.~\ref{app:budgets}). The shared cap grid is
\begin{equation}
\label{eq:setup_cap_grid}
\mathcal{C}=\left\{1,\frac{64}{255},\frac{32}{255},\frac{16}{255},
\frac{8}{255},\frac{4}{255},\frac{2}{255}\right\}.
\end{equation}
For every attack--model--budget frontier, $c^\star_{0.95}$ is independently
selected as the tightest cap retaining at least $95\%$ of its $c=1$ ASR.
Controls use operating points frozen at $B_{\mathrm{ref}}=4$ on CIFAR and
$20$ on ImageNet; robust points are selected independently and never retuned.
We also report $\rho\in\{0.90,0.99\}$ sensitivity analyses.

\paragraph{Controls, metrics, and reproducibility.}
Matched-realized-$\ell_1$ tests use APGD-L1 and SLIDE on $5{,}000$
CIFAR-10/ResNet-18 and $1{,}000$ ImageNet/ResNet-50 images, with shared sample
identity and $0.25\%$ relative matching tolerance. ImageNet perceptual tests
report LPIPS \cite{zhang2018unreasonable} and DISTS \cite{ding2020image} for all three
attacks at $c=1$, $c^\star_{0.95}$, and the next tighter cap. On $5{,}000$
CIFAR-10/ResNet-18 images, projection controls compare in-loop optimization
with post-hoc clipping, while stronger profiles increase APGD-L1 from one to
five restarts and FMN-L1 from $100$ to $500$ steps. A $700$-image ImageNet
audit evaluates APGD-L1 best-of-$k$, $k=1,\ldots,5$, and $500$-step FMN-L1 at
$c\in\{1,64/255\}$; official $\ell_1$ AutoAttack at $B=4,c=1$ on $1{,}000$
CIFAR-10/ResNet-18 images provides an external strength anchor. Full control
protocols appear in App.~\ref{app:controls}.

We record success, feasibility, realized $\ell_1$, $\ell_2$, and
$\ell_\infty$, top-$1\%$ and top-$5\%$ mass, effective support, cap
saturation, LPIPS, and DISTS. Structural summaries are medians over successful
feasible examples; matched analyses condition on capped-source success. We
report $95\%$ Wilson ASR intervals, $2{,}000$ deterministic paired-bootstrap
resamples for paired binary and median differences, and Spearman correlations
with sample-clustered resampling; frontiers are not smoothed. Statistical procedures,
sharding, and the full validation suite are detailed in
Apps.~\ref{app:metrics} and~\ref{app:implementation}.

\section{Results}
\label{sec:results}

% \section{Results}
% \label{sec:results}

We first report the perturbation--concentration frontiers on standard models,
then examine the corresponding behavior on adversarially trained models. We
finally use matched-realized-$\ell_1$ and perceptual analyses to characterize
the selected operating points. Optimization and implementation controls are
reported separately in Sec.~\ref{sec:ablation}.

\subsection{Main concentration results}
\label{sec:results_main}

Table~\ref{tab:main_concentration_standard} summarizes the primary operating
points for the standard models at $B=4$ for CIFAR-10 and CIFAR-100 and
$B=20$ for ImageNet.

Across all $18$ standard-model attack configurations, the $95\%$ retention
criterion relative to the native uncapped baseline is reached at an evaluated
cap of at most $32/255$. Median ASR retention is $97.02\%$, while median
peak magnitude decreases by $79.73\%$ and median effective support increases
by $176.56\%$.

Similar patterns are observed across datasets and architectures. For example,
APGD-L1 on ImageNet/ResNet-50 retains $95.45\%$ of its native uncapped ASR
at $4/255$, while reducing median $\ell_\infty$ by $96.69\%$ and increasing
effective support by $742.95\%$. CIFAR-100 also exhibits strong
redistribution, with APGD-L1 and FMN-L1 on WRN-28-10 selecting $2/255$.

\begin{figure}[t]
    \centering
    \includegraphics[width=\columnwidth]
    {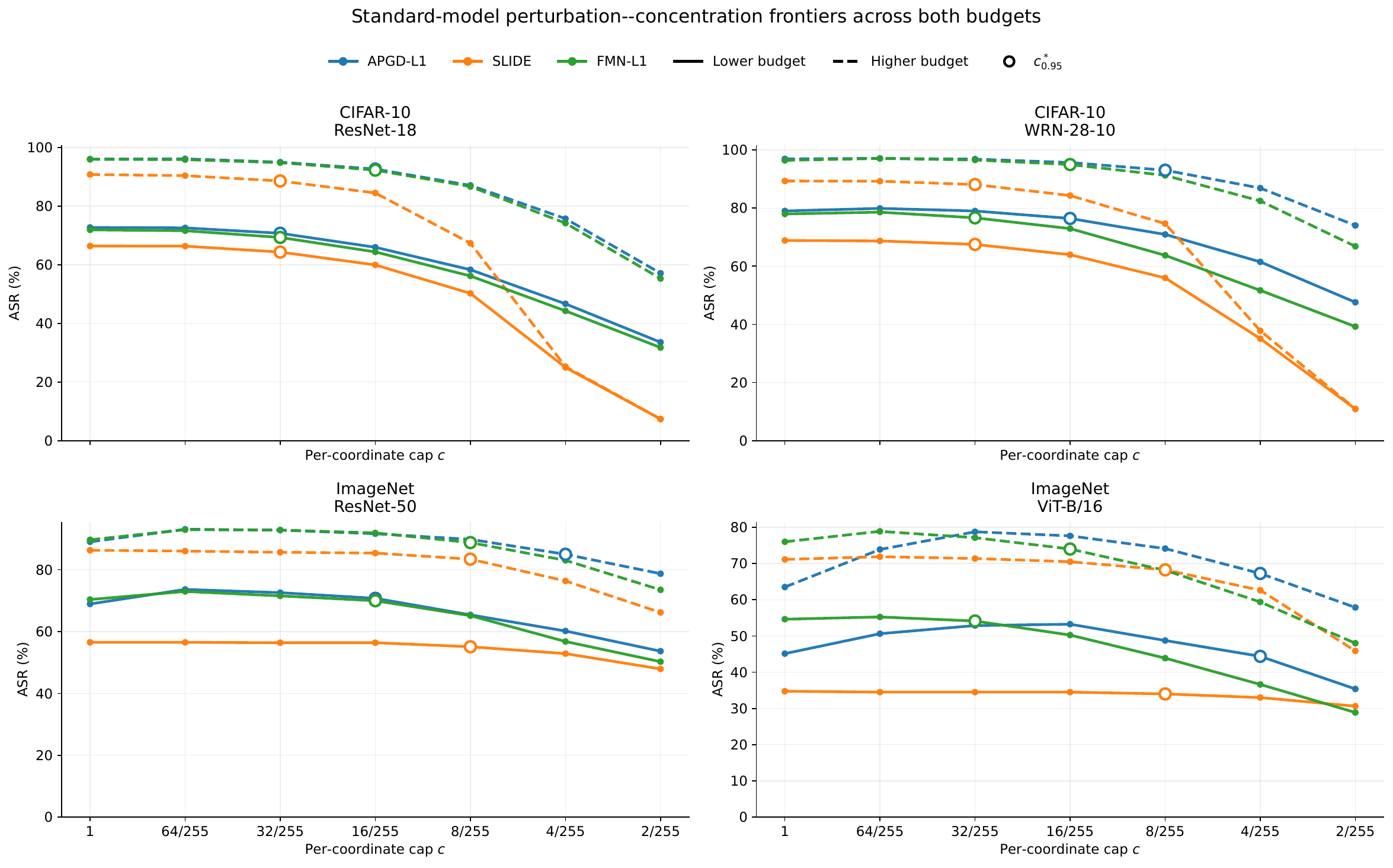}
    \caption{
    Standard-model perturbation--concentration frontiers across both global
    budgets. Panels show CIFAR-10/ResNet-18, CIFAR-10/WRN-28-10,
    ImageNet/ResNet-50, and ImageNet/ViT-B/16. Markers indicate the
    independently selected $c^\star_{0.95}$ values. CIFAR-100 results are
    included in Table~\ref{tab:main_concentration_standard}.
    }
    \label{fig:main_frontiers}
\end{figure}

\begin{figure*}[t]
    \centering
    \includegraphics[width=\textwidth]
    {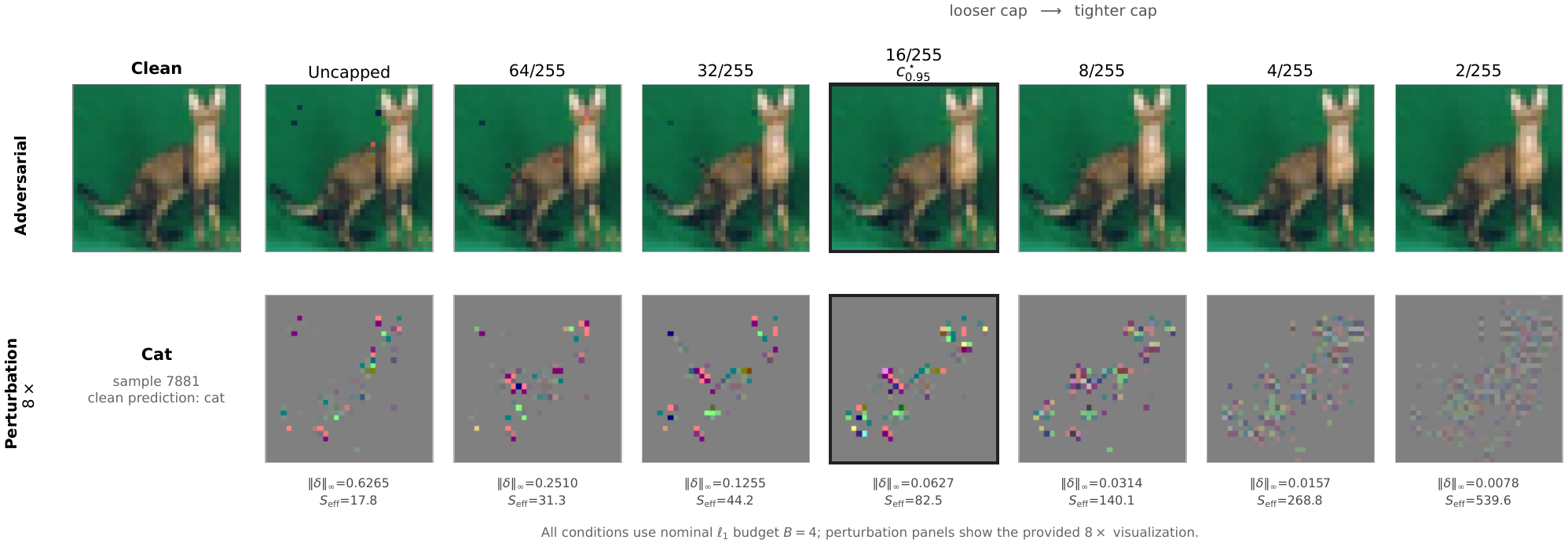}
    \caption{
    Qualitative perturbation progression under increasingly tight coordinate
    caps for a CIFAR-10 example at $B=4$. Tightening the cap progressively
    reduces peak magnitude and spreads the perturbation over larger effective
    support. The selected $c^\star_{0.95}=16/255$ operating point is
    highlighted; perturbations are shown at $8\times$ scale.
    }
    \label{fig:qualitative_cap_progression}
\end{figure*}

The measured frontiers occasionally exhibit finite-compute non-monotonicity.
For ImageNet/ViT-B/16, APGD-L1 achieves $63.50\%$ ASR at $c=1$,
$78.75\%$ at $32/255$, and $67.25\%$ at the selected $4/255$ cap.
Although the selected cap exceeds the native uncapped reference, it retains
only $85.4\%$ of the highest ASR observed across the evaluated cap grid.
Thus, the $95\%$ retention criterion quantifies effectiveness relative to
native implementations, not the strongest demonstrated feasible attack.
We report the unsmoothed frontiers and examine this finite-compute behavior
in Appendix~\ref{app:additional_frontier}.

\subsection{Adversarially trained models}
\label{sec:results_robust}

Table~\ref{tab:main_concentration_robust} summarizes the adversarially trained
models. Their measured concentration frontiers exhibit substantially more
heterogeneous behavior than those of standard classifiers.

The $\ell_1$-robust Croce--Hein model still admits substantial restriction:
APGD-L1 selects $32/255$ with $96.55\%$ retention. The multi-norm MSD model
is more concentration-sensitive under the evaluated attacks, with both
APGD-L1 and SLIDE selecting $64/255$.

The largest departure occurs for the $\ell_\infty$-trained XCiT models.
APGD-L1 requires the uncapped condition to satisfy the $95\%$ retention
criterion on both CIFAR-10 and ImageNet, whereas SLIDE selects $64/255$ and
$4/255$, respectively. These differences illustrate how the observed
concentration frontiers vary across model, training, and attack configurations.

\begin{figure}[t]
    \centering
    \includegraphics[width=\columnwidth]
    {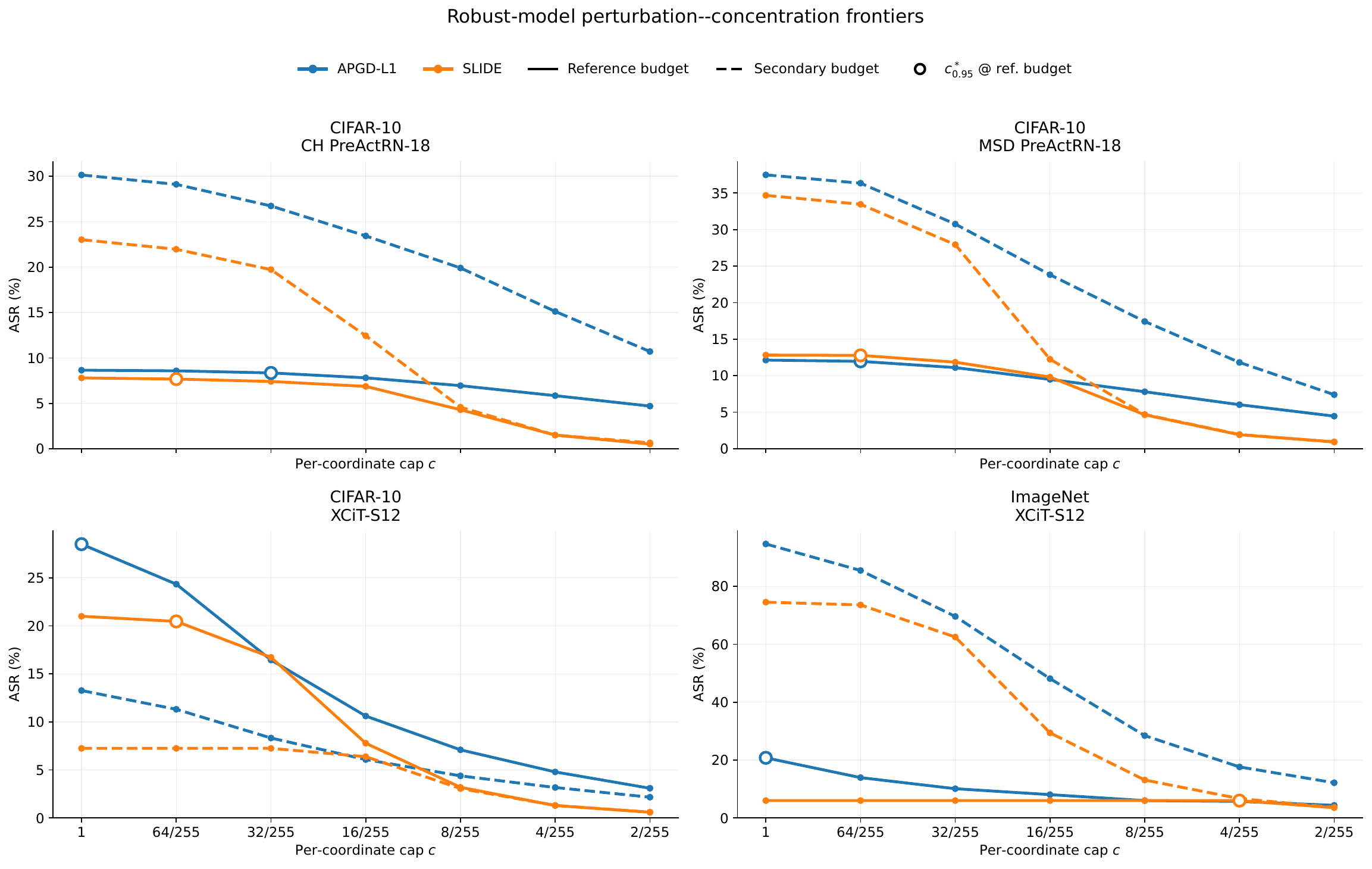}
    \caption{
    Perturbation--concentration frontiers for adversarially trained models.
    Solid curves denote the reference budget and dashed curves the
    corresponding secondary budget. Circled points indicate
    $c^\star_{0.95}$ at the reference budget.
    }
    \label{fig:robust_frontiers}
\end{figure}

\subsection{Matched realized $\ell_1$}
\label{sec:results_matched_l1}

The structural differences persist when capped perturbations are compared
with independently optimized uncapped perturbations at approximately matched
realized $\ell_1$ magnitude.

For APGD-L1, median $\ell_\infty$ decreases from $0.3060$ to $0.0627$ on
CIFAR-10/ResNet-18 and from $0.4569$ to $0.0157$ on
ImageNet/ResNet-50, while median effective support increases from $37.2$ to
$90.2$ and from $215.2$ to $1769.5$, respectively. The matched uncapped
attacks remain successful on $99.29\%$ and $96.95\%$ of the corresponding
capped-source-success populations.

Thus, the observed structural differences are not explained solely by reduced
realized $\ell_1$ use. Full structural comparisons are provided in
Appendix~\ref{app:matched_l1_results}.

\subsection{Perceptual-distance tradeoff}
\label{sec:results_perceptual}

On ImageNet/ResNet-50, the selected capped operating points also exhibit
lower learned perceptual distances. For APGD-L1, median LPIPS and DISTS
decrease by approximately $68.2\%$ and $61.6\%$ at $4/255$; for FMN-L1,
the corresponding reductions at $8/255$ are $38.1\%$ and $32.5\%$.

Paired common-success comparisons show the same direction. These results
support lower measured perceptual distortion under LPIPS and DISTS, not human
imperceptibility. Full numerical and visual comparisons are reported in
Appendix~\ref{app:perceptual_results}.

\paragraph{Overall finding.}
Across standard models, substantial empirical $\ell_1$ attack effectiveness
relative to native uncapped baselines is retained under restrictions on peak
coordinate magnitude, while successful perturbations become markedly more
distributed. Matched-$\ell_1$ analyses support redistribution beyond reduced
budget use, and perceptual measurements characterize additional differences
at selected ImageNet operating points. The robust-model results further
highlight the model- and attack-dependent nature of these observations.

\section{Ablations and Evaluation Audits}
\label{sec:ablation}

% \section{Ablation Studies}
% \label{sec:ablations}

\begin{table}[t]
    \centering
    \caption{
    Constraint-enforcement control (A04) on $4{,}750$ clean-correct
    CIFAR-10/ResNet-18 examples at $B=4$.
    $\Delta$ denotes direct constrained ASR minus post-hoc ASR.
    }
    \label{tab:projection_control}
    \setlength{\tabcolsep}{3.5pt}
    \scriptsize
    \begin{tabular}{lrrrr}
        \toprule
        Attack &
        $255c$ &
        Post-hoc &
        Direct &
        $\Delta$ ASR [95\% CI] \\
        \midrule
        APGD-L1 &
        16 & 46.72 & 92.21 &
        $+45.49\ [44.13,46.91]$ \\

        FMN-L1 &
        16 & 3.83 & 91.94 &
        $+88.11\ [87.20,88.99]$ \\
        \bottomrule
    \end{tabular}

\end{table}

The analyses in Sec.~\ref{sec:results} already examine two important aspects
of the observed differences: A01--A02 assess whether structural
redistribution persists after controlling for realized $\ell_1$
magnitude, and A03 measures the corresponding changes in learned
perceptual distance. We therefore focus here on the remaining controls:
constraint enforcement (A04), attack strength (A05), restart behavior (A06),
and the quality of the uncapped reference (A07). Complete protocols and
additional diagnostics are reported in Appendix~\ref{app:ablations}.

All ablations are performed after the primary $c^\star_{0.95}$ values are
frozen and therefore do not affect operating-point selection.

\subsection{Constraint enforcement and attack strength}
\label{sec:ablation_constraint_strength}

A04 tests whether the capped results could be reproduced simply by generating
an uncapped perturbation and clipping its largest coordinates afterward.
Table~\ref{tab:projection_control} compares post-hoc clipping with direct
optimization inside the same final feasible set on CIFAR-10/ResNet-18.

Direct constrained optimization achieves substantially higher ASR than
post-hoc clipping for both evaluated attacks. Since both final
perturbations satisfy the same declared constraints, the comparison
highlights the benefit of optimizing within the capped feasible set,
where perturbation mass can be redistributed, rather than truncating
a completed uncapped solution.

A05 examines whether the selected high-retention operating points remain
effective under stronger optimization. APGD-L1 is increased from one to
five restarts, while FMN-L1 is increased from $100$ to $500$ steps.

\begin{table}[t]
    \centering
    \caption{
    Optimization-strength control (A05) on CIFAR-10/ResNet-18 at $B=4$.
    ``Main'' and ``Strong'' denote the default and strengthened
    configurations. The selected operating points are frozen before this
    comparison.
    }
    \label{tab:strength_control}
    \setlength{\tabcolsep}{2.8pt}
    \scriptsize
    \begin{tabular}{llrrrr}
        \toprule
        Attack &
        Condition &
        $255c$ &
        Main &
        Strong &
        $\Delta$ ASR \\
        \midrule
        APGD-L1 & Uncapped  & 255 & 95.58 & 98.08 & +2.51 \\
                & $c^\star$ & 16  & 91.81 & 93.94 & +2.13 \\
                & Tighter   & 8   & 86.51 & 88.55 & +2.04 \\
        \addlinespace
        FMN-L1  & Uncapped  & 255 & 95.96 & 96.40 & +0.44 \\
                & $c^\star$ & 16  & 91.89 & 92.86 & +0.97 \\
                & Tighter   & 8   & 86.11 & 87.28 & +1.18 \\
        \bottomrule
    \end{tabular}
\end{table}

Additional compute improves both capped and uncapped conditions rather than
selectively recovering the conventional baseline. At the frozen operating
point, strengthened APGD-L1 retains $95.8\%$ of its strengthened uncapped
ASR and strengthened FMN-L1 retains $96.3\%$. Thus, within the evaluated
optimization profiles, stronger attacks change absolute ASR while
preserving the selected high-retention behavior.

\subsection{Finite-compute optimizer audit}
\label{sec:ablation_optimizer_audit}

A06 examines finite-compute non-monotonicity through a repeated-search
audit. On $700$ fixed clean-correct ImageNet/ResNet-50 examples at $B=20$,
we compare APGD-L1 at $c=1$ and $c=64/255$ over five independently
seeded runs.

\begin{figure}[t]
    \centering
    \includegraphics[width=\columnwidth]
    {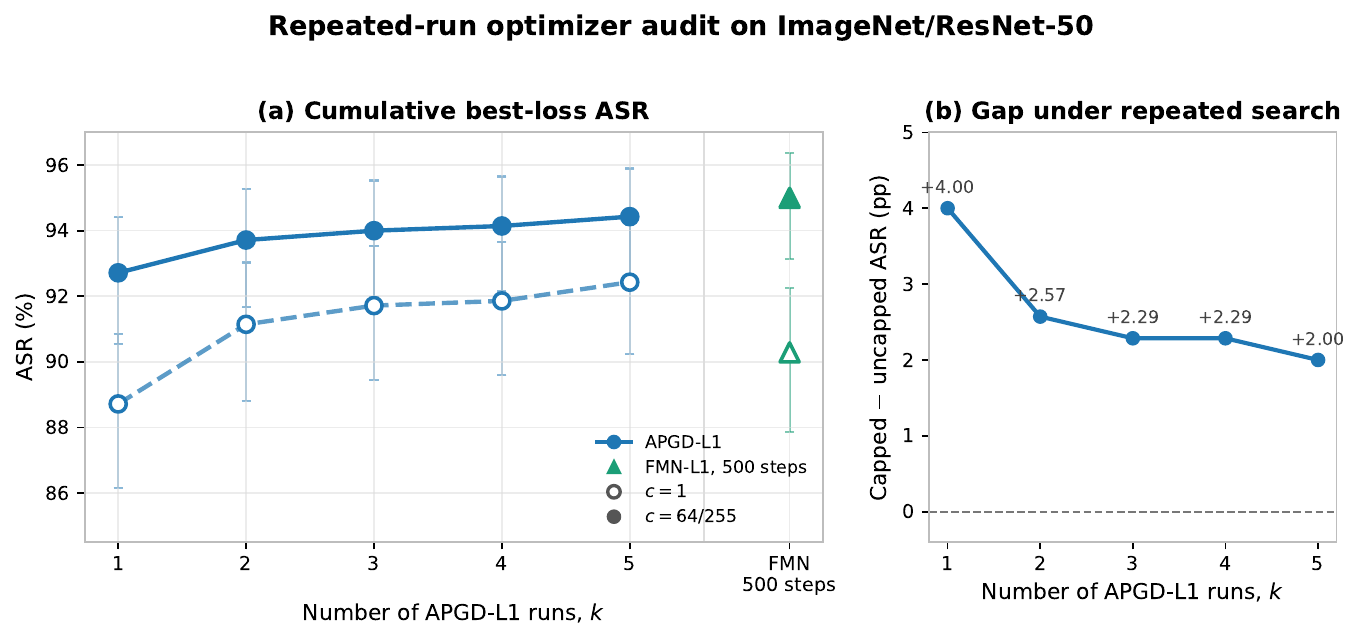}
    \caption{
    Repeated-run optimizer audit (A06). Cumulative best-of-$k$ APGD-L1 ASR,
    selecting the highest-loss candidate among the first $k$ runs, is shown
    for $c=1$ and $c=64/255$, with $95\%$ Wilson intervals.
    Markers show the corresponding $500$-step FMN-L1 references.
    Increasing independent search effort improves both conditions and narrows
    the capped-over-uncapped APGD gap.
    }
    \label{fig:restart_audit}
\end{figure}

APGD-L1 increases from $88.71\%$ to $92.43\%$ ASR between one and five
runs at $c=1$, and from $92.71\%$ to $94.43\%$ at $c=64/255$. The
capped-minus-uncapped gap therefore decreases from $4.00$ to $2.00$
percentage points. This narrowing is consistent with finite-compute
optimization contributing to the observed non-monotonicity, while the
remaining gap illustrates the dependence of empirical ASR on the search
procedure. Detailed best-of-$k$ values and broader frontier diagnostics
are provided in Appendix~\ref{app:optimizer_diagnostics}.

\subsection{Official AutoAttack reference}
\label{sec:ablation_autoattack}

A07 provides an external strength reference for the conventional uncapped
executions using the official unmodified untargeted $\ell_1$ AutoAttack
suite on $1{,}000$ fixed clean-correct CIFAR-10/ResNet-18 examples at $B=4$.

\begin{table}[t]
    \centering
    \caption{
    External uncapped attack-strength reference (A07).
    $\Delta$ denotes comparator ASR minus official AutoAttack ASR.
    }
    \label{tab:autoattack_reference}
    \setlength{\tabcolsep}{4.0pt}
    \scriptsize
    \begin{tabular}{lrr}
        \toprule
        Attack & ASR (\%) & $\Delta$ (pp) \\
        \midrule
        Official $\ell_1$ AutoAttack & 98.50 & --- \\
        APGD-L1                      & 96.20 & -2.30 \\
        FMN-L1                       & 95.70 & -2.80 \\
        \bottomrule
    \end{tabular}
\end{table}

The individual frontier attacks attain slightly lower ASR than the complete
AutoAttack ensemble, remaining within $2.8$ percentage points on the
matched population. This provides a reference for the strength of the
native uncapped executions in this setting. The official ensemble is
used only as an uncapped comparator because its standard form does not
enforce the additional coordinate cap.

Together with A01--A03, these controls provide complementary evidence
on realized budget use, perceptual measurements, constraint enforcement,
optimization strength, finite-compute restart behavior, and uncapped
reference quality within their respective evaluated settings. They
support the interpretation of the measured concentration frontiers
while distinguishing empirical attack behavior from guarantees about
optimal adversarial success. None of these controls changes the
operating points selected from the primary concentration frontiers.

\section{Conclusion}
\label{sec:conclusion}

We presented a concentration-aware evaluation of $\ell_1$ adversarial
attacks that distinguishes total perturbation magnitude from its
distribution across image coordinates. Across standard models on CIFAR-10,
CIFAR-100, and ImageNet, most attack effectiveness relative to native
uncapped baselines is retained under tighter per-coordinate caps, with
substantially reduced peak magnitude and increased effective support.
Matched-realized-$\ell_1$ comparisons confirm that these structural
differences persist beyond changes in budget utilization, while selected
ImageNet settings show lower LPIPS and DISTS. Adversarially trained models
exhibit more heterogeneous, attack-dependent sensitivity. Direct constrained
optimization also achieves up to $88$ percentage points higher attack
success than post-hoc clipping. Overall, perturbation concentration emerges
as a complementary dimension of empirical $\ell_1$ robustness evaluation,
highlighting the importance of constraint enforcement and finite-compute
optimization.

\bibliography{references}

\clearpage

\appendix
\setcounter{secnumdepth}{2}

\section{Projection Operator, Metric Definitions, and Protocol}
\label{app:method}

This appendix expands Sec.~\ref{sec:approach}. It derives the joint
projection used to realize the capped conditions, states the numerical checks
applied to it, gives the structural summary quantities in full, and records
the ordering of the evaluation protocol.

% ------------------------------------------------------------
\subsection{Joint projection: derivation and validation}
\label{app:projection_derivation}
% ------------------------------------------------------------

The projection of Eq.~\eqref{eq:approach_projection} solves
\begin{equation}
\label{eq:app_projection_primal}
\min_{z}\ \tfrac{1}{2}\|z-v\|_2^2
\quad\text{s.t.}\quad
\|z\|_1\leq B,
\quad
\ell_i\leq z_i\leq u_i,
\end{equation}
with $\ell_i=\max(-c,-x_i)$ and $u_i=\min(c,1-x_i)$. The feasible set is the
intersection of an $\ell_1$ ball with a box; it is convex and nonempty, since
$z=0$ satisfies both constraints for every $x\in[0,1]^D$. Because the
objective is strictly convex, the projected point exists and is unique.

Introducing a multiplier $\lambda\geq0$ for the $\ell_1$ constraint and
keeping the box constraints explicit, the Lagrangian separates across
coordinates. Each coordinate subproblem,
\begin{equation}
\label{eq:app_projection_coord}
\min_{\ell_i\leq z_i\leq u_i}
\tfrac{1}{2}(z_i-v_i)^2+\lambda|z_i|,
\end{equation}
is the proximal operator of $\lambda|\cdot|$ restricted to an interval, whose
solution is the soft threshold followed by a clip to that interval, giving
Eq.~\eqref{eq:approach_projection_solution}. The map
$\lambda\mapsto\|z(\lambda)\|_1$ is continuous, piecewise linear, and
non-increasing. Since $0\in[\ell_i,u_i]$ for every coordinate,
$z(\lambda)=0$ for all
\[
\lambda\geq\|v\|_\infty=\max_i |v_i|.
\]
Consequently, if the box projection at $\lambda=0$ already satisfies
$\|z(0)\|_1\leq B$, it is the solution. Otherwise, there exists
$\lambda^\star>0$ such that
$\|z(\lambda^\star)\|_1=B$, and a one-dimensional bisection search over
$[0,\|v\|_\infty]$ recovers such a multiplier to numerical precision.
Although the projected point is unique, uniqueness of the multiplier itself
is not required. Monotonicity of $\|z(\lambda)\|_1$ is sufficient to make the
search well posed; we use a fixed iteration count so that the operator is
deterministic.

Two properties are worth stating because they are what the capped conditions
rely on. The operator computes the joint Euclidean projection rather than
alternating between separate constraint projections, with the multiplier
resolved to numerical precision. It is also idempotent on feasible points:
applying it to an already-admissible perturbation leaves that perturbation
unchanged. This permits the projector to be inserted into an attack loop
without altering a feasible iterate when the constraint is inactive.

We validate the implementation numerically before the main runs. On random
inputs we verify that the returned point satisfies all three constraints of
Eq.~\eqref{eq:approach_threat_set} to solver tolerance, that the operator is
idempotent on feasible inputs, and that at $c=1$ it agrees with the standard
$\ell_1$-and-box projection, where the coordinate cap is mathematically
redundant. This last check establishes consistency of the joint projector
with the conventional uncapped feasible set. The empirical retention
reference remains the native uncapped implementation of each attack, as
specified in Sec.~\ref{sec:constraint_realization}.

% ------------------------------------------------------------
\subsection{Structural summary quantities}
\label{app:structural_defs}
% ------------------------------------------------------------

At the selected operating point we summarize the structural change relative
to the uncapped condition by the percentage reduction in median peak
magnitude,
\begin{equation}
\label{eq:app_linf_reduction}
\Delta_{\infty}
=
100
\left(
1-
\frac{\operatorname{median}\|\delta_{c^\star}\|_\infty}
{\operatorname{median}\|\delta_{1}\|_\infty}
\right),
\end{equation}
and the percentage increase in median effective support,
\begin{equation}
\label{eq:app_seff_increase}
\Delta_{S_{\mathrm{eff}}}
=
100
\left(
\frac{\operatorname{median}S_{\mathrm{eff}}(\delta_{c^\star})}
{\operatorname{median}S_{\mathrm{eff}}(\delta_{1})}
-1
\right).
\end{equation}
Both are computed over successful, feasible perturbations in the
corresponding condition. We summarize with medians rather than means because
the distributions of perturbation magnitudes and support measures are
strongly skewed: a small number of examples that saturate the cap or
concentrate almost all mass on a few coordinates would otherwise dominate the
comparison.

These two quantities are ratios of condition-level medians, not medians of
per-example ratios, and are therefore descriptive summaries of the two
conditions rather than paired per-example effects. Paired per-example
structural differences, where reported, are computed separately and stated as
such.

% ------------------------------------------------------------
\subsection{Ordered evaluation protocol}
\label{app:protocol}
% ------------------------------------------------------------

For each dataset, classifier, attack, and global budget, the evaluation
proceeds in the following order:
\begin{enumerate}
\itemsep0.15em
    \item evaluate the conventional uncapped attack at $c=1$ using its native
    implementation as the empirical reference;

    \item evaluate the same attack over the predefined sequence of tighter
    caps $c<1$, holding the global $\ell_1$ budget fixed and enforcing the
    joint constraint in-loop;

    \item construct the empirical perturbation--concentration frontier;

    \item select $c^\star_{0.95,\mathcal{A},f}(B)$ independently for that
    attack--model--budget configuration;

    \item compare uncapped and selected-cap perturbations using peak
    magnitude, top-mass concentration, effective support, and realized
    $\ell_1$;

    \item only then perform the downstream interpretive and validation
    analyses, including matched-realized-$\ell_1$, perceptual-distance,
    post-hoc clipping, optimization-strength, repeated-search, and
    external-reference controls.
\end{enumerate}
The ordering is not incidental. The operating point is determined solely from
ASR retention relative to the native uncapped reference on the predefined
frontier, before any downstream control is executed, so no control can
influence the selection it is meant to test. Steps~5 and~6 interpret and
stress-test the selected point; they never revise it. This is also why the
sensitivity analyses at $\rho\in\{0.90,0.99\}$ are performed only after the
primary selections are frozen and are never used to choose an alternative
operating point. Their results are reported separately in
Sec.~\ref{app:threshold_sensitivity}.

\section{Complete Experimental Setup}
\label{app:setup}

This appendix expands Sec.~\ref{sec:setup}. It records the evaluation
populations, model sources, attack instantiations, control protocols, and
validation procedures underlying the reported results. Where the main
text states a summary value, the corresponding subsection below describes
the procedure used to obtain it.

% ------------------------------------------------------------
\subsection{Datasets, manifests, and evaluation populations}
\label{app:data}
% ------------------------------------------------------------

The standard-model evaluation spans CIFAR-10 and CIFAR-100
\cite{krizhevsky2009learning} together with the ImageNet validation set
\cite{russakovsky2015imagenet}. CIFAR-10 and CIFAR-100 provide two
same-resolution settings with substantially different class granularity;
ImageNet provides a larger-scale, higher-resolution setting.

For CIFAR-10 and CIFAR-100, the frontier evaluation uses all $10{,}000$
images of the official test split in a fixed deterministic order. For
ImageNet, we construct a deterministic class-balanced manifest containing one
validation image from each of the $1{,}000$ classes; the image within each
class is selected once from the frozen experiment seed, and the resulting
manifest is shared across ImageNet models.

Attack success rate is computed only over examples classified correctly
before attack, so the clean-correct population may differ between models even
when they share a manifest. Within each dataset--model pair, all attacks,
budgets, and caps are evaluated on the same clean-correct population.
Throughout the paper, $N$ therefore denotes the clean-correct denominator
rather than the number of inputs initially loaded. An output counts as
successful only if it changes the predicted label and satisfies
\begin{equation}
\label{eq:app_feasibility}
\|\delta\|_1\leq B,
\qquad
\|\delta\|_\infty\leq c,
\qquad
x+\delta\in[0,1]^D,
\end{equation}
with feasibility verified numerically before aggregation.

% ------------------------------------------------------------
\subsection{Models, checkpoints, and preprocessing}
\label{app:models}
% ------------------------------------------------------------

\paragraph{Standard models.}
On both CIFAR-10 and CIFAR-100 we evaluate ResNet-18 \cite{he2016resnet} and
WideResNet-28-10 \cite{zagoruyko2016wide} using dataset-specific checkpoints;
the standard WideResNet models are loaded through RobustBench
\cite{croce2021robustbench}. On ImageNet we evaluate ResNet-50 and ViT-B/16
\cite{he2016resnet,dosovitskiy2021vit} using their published torchvision
ImageNet-1K weights. These six classifiers span two CIFAR label spaces and
both convolutional and transformer-based ImageNet architectures.

\paragraph{Adversarially trained models.}
We separately evaluate four adversarially trained classifiers spanning
different training threat models. On CIFAR-10 these are the Croce--Hein
PreActResNet-18 trained for $\ell_1$ robustness \cite{croce2021mind}, the MSD
PreActResNet-18 trained against the union of $\ell_\infty$, $\ell_2$, and
$\ell_1$ perturbations \cite{maini2020adversarial}, and the XCiT-S12 model
trained for $\ell_\infty$ robustness \cite{debenedetti2023light}. We
additionally evaluate the corresponding $\ell_\infty$-robust XCiT-S12 on
ImageNet.

The three CIFAR-10 robust frontiers start from the full $10{,}000$-image test
split, with audited clean-correct populations of $8{,}706$, $8{,}174$, and
$9{,}006$ examples for the Croce--Hein, MSD, and XCiT-S12 models
respectively. The ImageNet robust frontier uses the same deterministic
$1{,}000$-image class-balanced protocol and contains $731$ clean-correct
examples. These models are evaluated with their own complete concentration
frontiers and independently selected operating points; no $c^\star$ selected
on a standard model is transferred to a robust classifier.

\paragraph{Preprocessing.}
All attacks operate on final model inputs in the $[0,1]$ image domain, with
model-specific normalization applied inside the classifier and never folded
into the perturbation constraints. ImageNet models use the resize,
center-crop, and normalization associated with their published weights. All
budgets, caps, feasibility checks, and perturbation statistics are therefore
expressed in the same raw image space.

% ------------------------------------------------------------
\subsection{Attack instantiation}
\label{app:attacks}
% ------------------------------------------------------------

\paragraph{APGD-L1.}
Instantiated from the official AutoAttack implementation
\cite{croce2021mind}. At $c=1$ the unmodified upstream path is retained. For
$c<1$, only the constraint-projection operation is replaced by the joint
projection of Sec.~\ref{sec:constraint_realization}; the loss, gradient
construction, step adaptation, and candidate-selection logic are unchanged.

\paragraph{SLIDE.}
Our implementation follows the released optimization design
\cite{tramer2019adversarial}, including the margin objective, boundary
masking, Laplace initialization, and $99$th-percentile gradient
sparsification, and preserves the released relative step size by using
$0.255B$ at global budget $B$. The uncapped $c=1$ condition is evaluated
through independent executions using SLIDE's released native $\ell_1$
projection, while the capped conditions use the exact joint projector.
The native uncapped executions establish the conventional SLIDE reference
for retention, consistent with the other attack families. Accordingly,
the reported SLIDE frontiers characterize the capped implementation
relative to its native uncapped counterpart, with projection following
the respective feasibility-enforcement procedures.

\paragraph{FMN-L1.}
Instantiated from the authors-maintained Foolbox implementation
\cite{pintor2021fast}. Because FMN follows an adaptive minimum-norm
trajectory, the cap is imposed at the attack's current radius through the
corresponding joint feasibility projection. The attack is executed once per
cap and its candidate is subsequently evaluated at each applicable reporting
budget.

\paragraph{Profiles and randomness.}
The main standard-model profile uses $100$ optimization steps for all three
attacks; APGD-L1 and SLIDE use one restart and FMN-L1 one minimum-norm
trajectory. Corresponding cap conditions use deterministic common randomness:
for a fixed sample, attack, model, and budget the attack seed does not depend
on $c$, reducing stochastic variation between conditions. The robust-model
study evaluates APGD-L1 and SLIDE over the complete cap grid for all four
adversarially trained classifiers; these two attacks define the reported
robust frontiers. The additional optimization-strength, perceptual, and
external-reference controls are evaluated on the specific standard-model
settings described below.

% ------------------------------------------------------------
\subsection{Budgets, cap grid, and operating-point selection}
\label{app:budgets}
% ------------------------------------------------------------

Standard-model budgets are $B\in\{2,4\}$ on CIFAR-10 and CIFAR-100 and
$B\in\{10,20\}$ on ImageNet. Both budgets are first-class parts of the
frontier and budget-dependence analysis rather than a primary and a secondary
setting. The evaluated cap grid is given by Eq.~\eqref{eq:setup_cap_grid},
with $c=1$ the conventional image-domain reference.

The robust models are evaluated at $B\in\{4,12\}$ for the Croce--Hein and MSD
PreActResNet-18 models, $B\in\{2,4\}$ for CIFAR-10 XCiT-S12, and
$B\in\{20,255\}$ for ImageNet XCiT-S12. These ranges are chosen so that each
robust classifier is probed across a regime in which its own uncapped ASR is
neither saturated nor degenerate, which requires larger budgets than the
standard models at comparable resolution. The non-reference robust budgets
are consequently model-specific secondary probes: they are interpreted within
model and are never pooled as though they formed a common low-versus-high
budget experiment across classifiers.

For downstream controls and cross-regime comparisons we designate
$B_{\mathrm{ref}}=4$ on CIFAR and $B_{\mathrm{ref}}=20$ on ImageNet. For
every attack--model--budget frontier, $c^\star_{0.95}$ is computed
independently at the same budget, so the two budgets yield two independently
measured operating points rather than one cap reused across budgets. The
settings used by downstream controls are those selected at
$B_{\mathrm{ref}}$, frozen before the matched-realized-$\ell_1$, perceptual,
projection, optimization-strength, and qualitative analyses are executed.
Robust operating points are selected independently from the completed robust
APGD-L1 and SLIDE frontiers at the corresponding reference budget and are
never retuned by a later stage. The complete frontier is always retained and
reported; $c^\star$ serves only as a compact summary and as the condition
used in subsequent structural analyses. We additionally recompute the summary
criterion at $\rho\in\{0.90,0.99\}$ as a threshold-sensitivity analysis,
which affects no downstream experiment. The resulting retention-threshold
and cross-budget operating-point analyses are reported in
Secs.~\ref{app:threshold_sensitivity} and~\ref{app:budget_sensitivity},
respectively.

% ------------------------------------------------------------
\subsection{Controlled evaluations}
\label{app:controls}
% ------------------------------------------------------------

\paragraph{Matched realized $\ell_1$.}
Performed on CIFAR-10/ResNet-18 using a fixed $5{,}000$-image subset and on
ImageNet/ResNet-50 using the complete $1{,}000$-image manifest, for APGD-L1
and SLIDE, the two attacks for which an example-specific uncapped budget can
be assigned directly. For every successful capped source perturbation, its
realized $\|\delta\|_1$ becomes the budget of an independently optimized
uncapped execution on the same sample, so sample identity is preserved across
conditions. The matched execution must reproduce the source magnitude within
a relative tolerance of $0.25\%$, and we report the residual relative
norm-matching error rather than relying on the requested budget alone.
Because the analysis begins from successful capped source perturbations, the
comparison is explicitly conditioned on capped-source success, and the
success rate of the matched uncapped execution is a diagnostic on that
conditioned population rather than a full-dataset ASR.

\paragraph{Perceptual distance.}
Computed on ImageNet/ResNet-50 for APGD-L1, SLIDE, and FMN-L1 at the uncapped
reference, the selected $c^\star_{0.95}$ condition at $B_{\mathrm{ref}}$, and
the next tighter evaluated cap, using LPIPS \cite{zhang2018unreasonable} and
DISTS \cite{ding2020image}. We report both condition-wise summaries and paired
common-success analyses, the latter restricting the comparison to samples on
which the relevant conditions are jointly successful in order to reduce
confounding from a changing success population. Both quantities are treated
as learned perceptual-distance measures and are used to characterize how
measured perceptual distortion varies across the evaluated cap conditions.

\paragraph{Constraint enforcement and optimization strength.}
The constraint-enforcement control compares direct optimization under the
exact joint projection against post-hoc clipping on a fixed $5{,}000$-image
CIFAR-10/ResNet-18 population. In the post-hoc condition an uncapped
perturbation is optimized first and its coordinates are subsequently
truncated to the selected cap. The comparison tests whether direct
optimization within the capped feasible set achieves higher attack success
than modifying a completed uncapped solution. The optimization-strength
control uses the same dataset--model setting: APGD-L1 retains $100$ steps and
increases from one to five restarts, and FMN-L1 increases from $100$ to $500$
steps.

\paragraph{Repeated-run audit and external anchor.}
The repeated-run optimizer audit uses $700$ fixed clean-correct
ImageNet/ResNet-50 examples selected deterministically from the corresponding
main-run population, evaluating APGD-L1 through cumulative best-of-$k$
results for $k\in\{1,\ldots,5\}$, selecting the highest-loss candidate
among the first $k$ runs, together with an independent $500$-step
FMN-L1 reference at $c\in\{1,64/255\}$. As an external reference for the
strength of the conventional uncapped evaluation, we run the official
untargeted $\ell_1$ AutoAttack suite at $B=4$ and $c=1$ on $1{,}000$ fixed
clean-correct CIFAR-10/ResNet-18 examples. Because both audits are performed
on populations preselected to be clean-correct, the reported quantity is ASR
on those fixed examples rather than full-dataset robust accuracy, and the
sample IDs are fixed deterministically before either audit is executed.
Neither control modifies the operating points selected from the primary
sweeps.

\paragraph{Qualitative examples.}
Qualitative examples are selected only after all quantitative operating
points are frozen, and are used solely to visualize clean images, capped and
uncapped adversarial examples, and their perturbations at selected settings.
They contribute to no cap selection, attack tuning, or aggregate metric.

% ------------------------------------------------------------
\subsection{Metrics and statistical analysis}
\label{app:metrics}
% ------------------------------------------------------------

For every attack output we record attack success, feasibility, realized
$\ell_1$, $\ell_2$, and $\ell_\infty$ magnitudes, top-$1\%$ and top-$5\%$
mass fractions, effective support, and cap saturation, together with LPIPS and
DISTS where applicable. Unless stated otherwise, perturbation-structure
summaries are reported as medians over successful, feasible perturbations;
medians are used because the distributions of perturbation magnitudes and
support measures are strongly skewed. At the selected operating point we
additionally report ASR retention, the relative reduction in median
$\ell_\infty$, and the relative increase in median effective support.

Paired analyses preserve sample identity and state any additional
conditioning explicitly; in particular, matched-realized-$\ell_1$ analyses are
conditioned on capped-source success whereas ordinary ASR estimates use the
full clean-correct population. We report $95\%$ Wilson confidence intervals
for individual ASR estimates. Differences between paired binary outcomes and
median paired structural changes use $2{,}000$ deterministic paired-bootstrap
resamples. Supplementary correlation analyses use Spearman correlation with
sample-level clustered resampling; the corresponding descriptive estimates
are reported in Sec.~\ref{app:metric_correlations}. These correlations are
used only to characterize associations among measured quantities and are not
interpreted causally. Empirical frontiers are always reported without
monotonic smoothing.

% ------------------------------------------------------------
\subsection{Implementation, sharding, and validation}
\label{app:implementation}
% ------------------------------------------------------------

Experiments are implemented in PyTorch and executed on NVIDIA T4 GPUs. Runs
are deterministically partitioned into disjoint external shards where required
by dataset size, model memory, or runtime; the completed robust executions
R01--R04 used one, one, three, and two external shards respectively. Each
worker holds an independent model replica and writes a disjoint output shard.
Sharding changes only execution partitioning: sample identity, attack
configuration, and randomness are derived independently of worker assignment.

The exact joint projection uses $32$ bisection iterations. All data sampling,
manifest construction, attack randomness, and bootstrap procedures are derived
from the frozen base seed $20260822$, and corresponding cap conditions use
common randomness wherever the attack formulation permits it.

Before aggregation we verify configuration hashes, manifest hashes,
worker and shard completeness, sample-set consistency, condition uniqueness,
expected row counts, attack feasibility, and the absence of duplicate runs.
Results from separate execution sessions are merged only after these checks
pass. All final figures and tables are generated from the validated scalar
outputs and the frozen operating-point records; no attack is rerun during
figure or table generation.

\section{Additional Structural and Perceptual Results}
\label{app:additional_results}

This appendix expands the matched-realized-$\ell_1$ and perceptual analyses
summarized in Secs.~\ref{sec:results_matched_l1} and
\ref{sec:results_perceptual}. These analyses use operating points selected
from the primary concentration frontiers and do not influence
$c^\star_{0.95}$ selection.

% ------------------------------------------------------------
\subsection{Matched-realized-$\ell_1$ structural comparison}
\label{app:matched_l1_results}
% ------------------------------------------------------------

The matched-realized-$\ell_1$ control compares each successful capped
perturbation with an independently optimized uncapped perturbation on the same
sample using approximately the same realized $\ell_1$ magnitude. This
separates a genuine change in perturbation allocation from the simpler
possibility that capped attacks merely use less total perturbation.

\begin{figure}[t]
    \centering
    \includegraphics[width=\columnwidth]
    {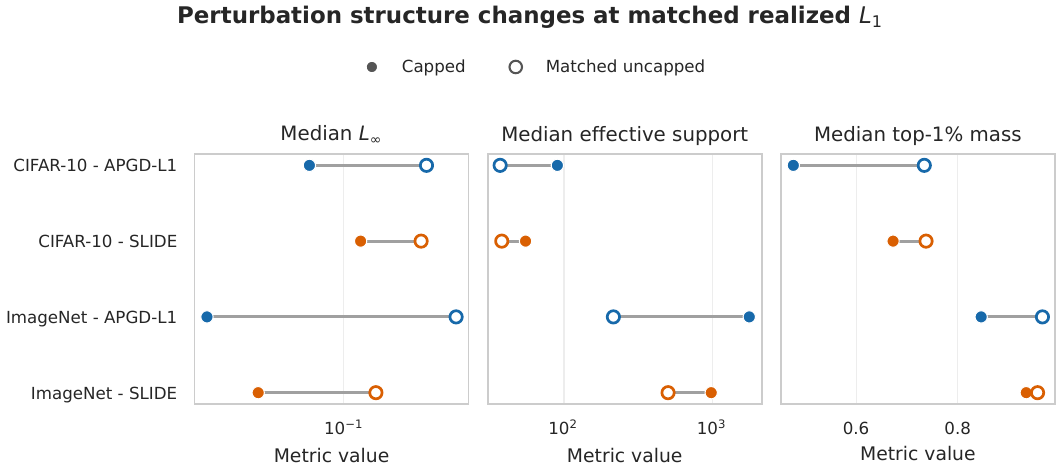}
    \caption{
    Perturbation structure under matched realized $\ell_1$. Filled markers
    denote capped perturbations and open markers independently optimized
    uncapped perturbations with approximately matched realized $\ell_1$.
    Capped solutions exhibit lower peak magnitude, larger effective support,
    and reduced concentration in the largest coordinates.
    }
    \label{fig:app_matched_l1_structure}
\end{figure}

Figure~\ref{fig:app_matched_l1_structure} shows the resulting structural
differences. Across both CIFAR-10 and ImageNet, capped perturbations occupy a
lower-$\ell_\infty$, higher-effective-support regime than their matched
uncapped counterparts. Table~\ref{tab:app_matched_l1_full} gives the complete
condition-level structural summaries for all four completed matching
experiments.

For APGD-L1 on CIFAR-10/ResNet-18, median $\ell_\infty$ decreases from
$0.3060$ in the matched uncapped condition to $0.0627$ under the cap, while
median effective support increases from $37.2$ to $90.2$. The independently
optimized matched uncapped attack succeeds on $99.29\%$ of the
capped-source-success population.

On ImageNet/ResNet-50, median $\ell_\infty$ decreases from $0.4569$ to
$0.0157$, while median effective support increases from $215.2$ to
$1769.5$. The matched uncapped attack succeeds on $96.95\%$ of the
corresponding conditioned population.

The same qualitative separation is present for SLIDE. On CIFAR-10,
median $\ell_\infty$ changes from $0.2846$ in the matched uncapped condition
to $0.1255$ under the cap, while effective support changes from $38.1$ to
$55.1$. On ImageNet, the corresponding medians are $0.1543$ versus $0.0314$
for $\ell_\infty$ and $502.3$ versus $980.5$ for effective support. Matched
uncapped SLIDE succeeds on $99.17\%$ and $99.84\%$ of the conditioned
CIFAR-10 and ImageNet populations, respectively.

\begin{table*}[t]
    \centering
    \caption{
    Complete matched-realized-$\ell_1$ structural summaries. All rows are
    conditioned on success of the capped source perturbation. ``C/U'' denotes
    capped source / independently optimized matched uncapped. Matching-error
    columns report the relative error between the realized $\ell_1$ magnitudes
    of the paired perturbations.
    }
    \label{tab:app_matched_l1_full}
    \small
    \setlength{\tabcolsep}{4pt}
    \begin{tabular}{llrrrrrrrr}
        \toprule
        Data & Attack & $N_{\mathrm{pairs}}$
        & Matched ASR
        & Med.\ err.
        & P95 err.
        & $\ell_\infty$ C/U
        & $M_{1\%}$ C/U
        & $M_{5\%}$ C/U
        & $S_{\mathrm{eff}}$ C/U \\
        \midrule
        CIFAR-10
        & APGD-L1
        & 4367
        & 99.29
        & $2.38{\times}10^{-7}$
        & $1.31{\times}10^{-6}$
        & .0627/.3060
        & .4765/.7342
        & .9865/1.0000
        & 90.2/37.2 \\

        CIFAR-10
        & SLIDE
        & 4194
        & 99.17
        & $1.19{\times}10^{-7}$
        & $2.38{\times}10^{-7}$
        & .1255/.2846
        & .6728/.7377
        & 1.0000/1.0000
        & 55.1/38.1 \\

        ImageNet
        & APGD-L1
        & 656
        & 96.95
        & $8.58{\times}10^{-7}$
        & $9.25{\times}10^{-6}$
        & .0157/.4569
        & .8462/.9668
        & 1.0000/1.0000
        & 1769.5/215.2 \\

        ImageNet
        & SLIDE
        & 640
        & 99.84
        & $1.91{\times}10^{-7}$
        & $1.24{\times}10^{-6}$
        & .0314/.1543
        & .9349/.9572
        & 1.0000/1.0000
        & 980.5/502.3 \\
        \bottomrule
    \end{tabular}
\end{table*}

Across the completed matching controls, median relative $\ell_1$ matching
error is below $10^{-6}$, with the largest $95$th-percentile relative error
equal to $9.3\times10^{-6}$. The observed structural separation is therefore
not attributable to a meaningful mismatch in realized total perturbation
magnitude.

Because the matching procedure preserves sample identity, we additionally
compute paired structural differences and deterministic $95\%$ bootstrap
intervals. These are reported in Table~\ref{tab:app_matched_l1_paired}.
For peak magnitude and top-mass concentration, positive differences denote
larger values for the matched uncapped perturbation; for effective support,
positive differences denote larger values for the capped perturbation.
The intervals therefore provide a paired complement to the condition-level
medians above.

\begin{table*}[t]
    \centering
    \caption{
    Paired structural differences in the matched-realized-$\ell_1$ controls,
    with $95\%$ deterministic paired-bootstrap intervals. For
    $\ell_\infty$, $M_{1\%}$, and $M_{5\%}$, differences are
    matched-uncapped minus capped; for $S_{\mathrm{eff}}$, differences are
    capped minus matched-uncapped. Positive values therefore indicate broader,
    lower-peak allocation under the cap.
    }
    \label{tab:app_matched_l1_paired}
    \small
    \setlength{\tabcolsep}{5pt}
    \begin{tabular}{llrrrr}
        \toprule
        Data & Attack
        & $\Delta\ell_\infty$ [95\% CI]
        & $\Delta M_{1\%}$ [95\% CI]
        & $\Delta M_{5\%}$ [95\% CI]
        & $\Delta S_{\mathrm{eff}}$ [95\% CI] \\
        \midrule

        CIFAR-10
        & APGD-L1
        & .2433 [.2392,.2483]
        & .2620 [.2585,.2662]
        & .0010 [.0004,.0019]
        & 52.85 [52.47,53.28] \\

        CIFAR-10
        & SLIDE
        & .1592 [.1555,.1627]
        & .0644 [.0626,.0667]
        & .0000 [.0000,.0000]
        & 16.68 [16.45,16.99] \\

        ImageNet
        & APGD-L1
        & .4413 [.4284,.4524]
        & .1064 [.1020,.1104]
        & .0000 [.0000,.0000]
        & 1536.43 [1515.61,1555.61] \\

        ImageNet
        & SLIDE
        & .1230 [.1199,.1256]
        & .0204 [.0193,.0215]
        & .0000 [.0000,.0000]
        & 477.67 [471.15,482.81] \\
        \bottomrule
    \end{tabular}
\end{table*}

The top-$5\%$ mass statistic is close to saturation in several matched
conditions, particularly on ImageNet, and consequently provides less
separation than peak magnitude, top-$1\%$ mass, or effective support. We
therefore treat these statistics as complementary rather than requiring every
concentration measure to change by the same amount.

% ------------------------------------------------------------
\subsection{Perceptual-distance tradeoff}
\label{app:perceptual_results}
% ------------------------------------------------------------

We additionally evaluate whether the structural redistribution at the selected
operating points is accompanied by changes in learned perceptual-distance
measures on ImageNet/ResNet-50.

For APGD-L1, moving from the uncapped condition to $c^\star=4/255$ changes
ASR from $89.43\%$ to $85.51\%$. Over the same comparison, median LPIPS
decreases from $0.0155$ to $0.0049$ and median DISTS from $0.0194$ to
$0.0074$, corresponding to reductions of approximately $68.2\%$ and
$61.6\%$.

For SLIDE, moving from the uncapped condition to the selected
$c^\star=8/255$ changes ASR from $86.55\%$ to $83.68\%$. Median LPIPS
decreases from $0.00975$ to $0.00725$ and median DISTS from $0.01264$ to
$0.01037$, corresponding to reductions of approximately $25.7\%$ and
$17.9\%$, respectively.

For FMN-L1, the selected $c^\star=8/255$ condition changes ASR from
$90.08\%$ to $88.64\%$, a decrease of $1.44$ percentage points, while
median LPIPS decreases from $0.00410$ to $0.00254$ and median DISTS from
$0.00653$ to $0.00441$. These correspond to reductions of approximately
$38.1\%$ and $32.5\%$, respectively.

\begin{figure}[t]
    \centering
    \includegraphics[width=\columnwidth]
    {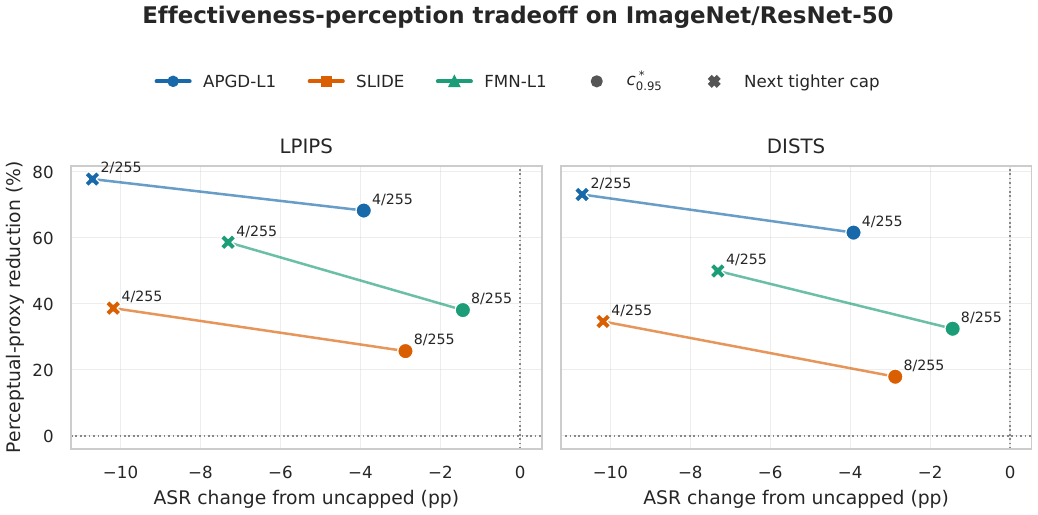}
    \caption{
    Effectiveness--perceptual-distance tradeoff on ImageNet/ResNet-50.
    Circles denote the selected $c^\star_{0.95}$ and crosses the next tighter
    evaluated cap. The horizontal axis reports ASR change relative to the
    uncapped condition, while the vertical axis reports reduction in LPIPS or
    DISTS.
    }
    \label{fig:app_perceptual_tradeoff}
\end{figure}

Figure~\ref{fig:app_perceptual_tradeoff} places these changes in the full
effectiveness--perceptual-distance tradeoff, together with the next tighter
evaluated caps.

Because the successful population changes with the cap, we also perform
paired common-success comparisons between each capped condition and its
uncapped reference. Table~\ref{tab:app_perceptual_paired} reports these
paired results for all three attacks at both the selected operating point and
the next tighter evaluated cap. ASR differences are computed on the fixed
$N=766$ clean-correct evaluation population, while the paired LPIPS and DISTS
differences are computed on examples successful under both conditions.
Negative perceptual differences indicate lower measured distance under the
capped condition.

\begin{table*}[t]
    \centering
    \caption{
    Paired ImageNet/ResNet-50 perceptual comparisons against the native
    uncapped reference. $\Delta$ASR is target minus uncapped in percentage
    points. $\Delta$LPIPS and $\Delta$DISTS are paired median differences
    (target minus uncapped) on the corresponding common-success population.
    Brackets give $95\%$ deterministic paired-bootstrap intervals.
    }
    \label{tab:app_perceptual_paired}
    \small
    \setlength{\tabcolsep}{6pt}
    \begin{tabular}{llrrrr}
        \toprule
        Attack
        & Target
        & $255c$
        & $\Delta$ASR (pp) [95\% CI]
        & $\Delta$LPIPS [95\% CI]
        & $\Delta$DISTS [95\% CI] \\
        \midrule

        APGD-L1
        & $c^\star$
        & 4
        & $-3.92$ [$-5.87$,$-1.96$]
        & $-0.01048$ [$-0.01145$,$-0.00970$]
        & $-0.01104$ [$-0.01213$,$-0.01017$] \\

        APGD-L1
        & tighter
        & 2
        & $-10.70$ [$-13.19$,$-8.36$]
        & $-0.01264$ [$-0.01358$,$-0.01163$]
        & $-0.01351$ [$-0.01438$,$-0.01272$] \\

        SLIDE
        & $c^\star$
        & 8
        & $-2.87$ [$-4.18$,$-1.83$]
        & $-0.001895$ [$-0.002001$,$-0.001702$]
        & $-0.002205$ [$-0.002412$,$-0.002000$] \\

        SLIDE
        & tighter
        & 4
        & $-10.18$ [$-12.40$,$-8.22$]
        & $-0.003466$ [$-0.003785$,$-0.003252$]
        & $-0.004441$ [$-0.004701$,$-0.004201$] \\

        FMN-L1
        & $c^\star$
        & 8
        & $-1.44$ [$-3.52$,$+0.53$]
        & $-0.001209$ [$-0.001360$,$-0.001013$]
        & $-0.001406$ [$-0.001592$,$-0.001215$] \\

        FMN-L1
        & tighter
        & 4
        & $-7.31$ [$-9.79$,$-4.83$]
        & $-0.001592$ [$-0.001807$,$-0.001394$]
        & $-0.002181$ [$-0.002438$,$-0.001996$] \\
        \bottomrule
    \end{tabular}
\end{table*}

The paired common-success results follow the same direction as the
condition-wise summaries. At each selected operating point, the paired
bootstrap intervals for both LPIPS and DISTS lie below zero. The same is true
at the next tighter evaluated cap. This reduces the possibility that the
condition-wise perceptual differences arise only because progressively
different examples remain successfully attacked.

The condition-wise medians and paired differences answer related but distinct
questions. Condition-wise summaries describe the successful feasible
perturbations produced under each condition separately, whereas the paired
analysis restricts the comparison to examples jointly successful under the
relevant uncapped and capped conditions. The paired median differences
therefore need not equal the simple difference between the two
condition-level medians.

Taken together, the condition-wise and paired analyses provide consistent
evidence that the selected capped conditions reduce learned perceptual
distance while retaining high attack effectiveness. The effect is observed
under both LPIPS and DISTS and persists when the comparison is restricted to
examples that are successfully attacked under both the capped and uncapped
conditions. In this measured sense, the resulting adversarial examples are
closer to the corresponding clean images: the cap restricts the largest
coordinate-wise modifications, for example to $8/255$ at the selected
operating point for SLIDE and FMN-L1, while LPIPS and DISTS also decrease.
Although these measurements do not directly establish that the perturbations
are less detectable by human observers, reducing large local pixel changes
and moving the adversarial examples closer to the clean images under learned
perceptual metrics is consistent with perturbations having lower visual
salience.

\section{Complete Ablation and Control Results}
\label{app:ablations}

This appendix expands Sec.~\ref{sec:ablation} and records the complete
control suite used to examine alternative explanations for the primary
concentration results. The controls are denoted A01--A07. None participates
in selecting the primary $c^\star_{0.95}$ operating points.

A01--A02 evaluate matched realized $\ell_1$, A03 evaluates learned
perceptual distance, A04 compares direct constraint enforcement with post-hoc
clipping, A05 strengthens the main attacks, A06 audits restart-dependent
finite-compute behavior, and A07 compares the uncapped frontier attacks with
official $\ell_1$ AutoAttack.

% ------------------------------------------------------------
\subsection{Matched-realized-$\ell_1$ controls: A01--A02}
\label{app:matched_l1}
% ------------------------------------------------------------

A01 and A02 examine whether lower peak magnitude and larger effective support
under the selected caps can be explained merely by the capped attack using
less total $\ell_1$ magnitude.

For every successful capped source perturbation, we independently rerun the
corresponding uncapped attack on the same sample with the source perturbation's
realized $\ell_1$ magnitude used as its per-example budget. The resulting
comparison therefore preserves sample identity and approximately matches
total perturbation magnitude across capped and native uncapped executions.

The CIFAR-10 control uses ResNet-18 and a fixed $5{,}000$-image subset. The
ImageNet control uses ResNet-50 and the complete $1{,}000$-image manifest.
The analysis is conditioned on success of the capped source perturbation.

For APGD-L1 on CIFAR-10/ResNet-18, median $\ell_\infty$ is $0.0627$ for
the capped source and $0.3060$ for the matched uncapped perturbation, while
median effective support is $90.2$ and $37.2$, respectively. The matched
uncapped attack succeeds on $99.29\%$ of the conditioned population.

On ImageNet/ResNet-50, median $\ell_\infty$ is $0.0157$ for the capped
source and $0.4569$ for the matched uncapped condition, while effective
support is $1769.5$ and $215.2$. The independently optimized matched
uncapped attack succeeds on $96.95\%$ of the conditioned population.

Across the completed controls, median relative $\ell_1$ matching error is
below $10^{-6}$ and the largest $95$th-percentile error is
$9.3\times10^{-6}$. The structural separation is therefore not explained by
a meaningful mismatch in realized total perturbation magnitude.

The complete APGD-L1 and SLIDE condition-level summaries, top-mass
concentration measurements, and paired structural differences with bootstrap
intervals are reported in Sec.~\ref{app:matched_l1_results}. Those results
show that the matched-$\ell_1$ separation is not specific to the APGD-L1
examples summarized here.

% ------------------------------------------------------------
\subsection{Perceptual-distance control: A03}
\label{app:perceptual}
% ------------------------------------------------------------

A03 evaluates whether the structural redistribution is accompanied by a
change in learned image-space perceptual distance. The analysis uses
ImageNet/ResNet-50 and compares the native uncapped condition, the selected
$c^\star_{0.95}$ condition, and the next tighter evaluated cap.

For APGD-L1, moving from the uncapped condition to $c^\star=4/255$ changes
ASR from $89.43\%$ to $85.51\%$, while median LPIPS decreases from
$0.0155$ to $0.0049$ and median DISTS from $0.0194$ to $0.0074$.

For FMN-L1, moving to $c^\star=8/255$ changes ASR by only $1.44$
percentage points while reducing median LPIPS from $0.00410$ to $0.00254$
and median DISTS from $0.00653$ to $0.00441$.

The paired common-success analyses show the same direction and are used to
reduce confounding from changes in the set of successful examples. Complete
paired results for APGD-L1, SLIDE, and FMN-L1 at both $c^\star_{0.95}$ and
the next tighter cap are reported in Sec.~\ref{app:perceptual_results}.
These results characterize differences in learned perceptual-distance
metrics across the evaluated conditions and do not establish human
imperceptibility, visibility thresholds, or perceptual preference.

% ------------------------------------------------------------
\subsection{Constraint-enforcement control: A04}
\label{app:projection_control}
% ------------------------------------------------------------

A04 compares direct optimization under the joint feasible set with post-hoc
clipping. Both final perturbations are required to satisfy the same global
$\ell_1$, coordinate-cap, and image-domain constraints.

\begin{table}[t]
    \centering
    \caption{
    Complete constraint-enforcement control on
    CIFAR-10/ResNet-18 ($N=4{,}750$, $B=4$).
    Confidence intervals are deterministic paired-bootstrap intervals.
    }
    \label{tab:app_projection_control}
    \setlength{\tabcolsep}{3.5pt}
    \scriptsize
    \begin{tabular}{lrrrr}
        \toprule
        Attack &
        $255c$ &
        Post-hoc &
        Direct &
        $\Delta$ ASR [95\% CI] \\
        \midrule
        APGD-L1 &
        16 & 46.72 & 92.21 &
        $+45.49\ [44.13,46.91]$ \\

        FMN-L1 &
        16 & 3.83 & 91.94 &
        $+88.11\ [87.20,88.99]$ \\
        \bottomrule
    \end{tabular}
\end{table}

The large gap demonstrates the substantial benefit of optimizing directly
within the joint feasible region under the evaluated conditions. Truncating
an uncapped solution does not allow removed mass to be reassigned to other
coordinates and therefore produces much lower attack success in these
comparisons. The reported control focuses on APGD-L1 and FMN-L1.

% ------------------------------------------------------------
\subsection{Optimization-strength control: A05}
\label{app:strength_control}
% ------------------------------------------------------------

A05 increases attack effort without changing the feasible set or operating
points. APGD-L1 uses five restarts instead of one, with $100$ steps per
restart. FMN-L1 uses $500$ steps instead of $100$.

\begin{table}[t]
    \centering
    \caption{
    Complete optimization-strength control on
    CIFAR-10/ResNet-18 ($N=4{,}750$, $B=4$).
    }
    \label{tab:app_strength_control}
    \setlength{\tabcolsep}{2.5pt}
    \scriptsize
    \begin{tabular}{llrrrrr}
        \toprule
        Attack &
        Condition &
        $255c$ &
        Main &
        Strong &
        $\Delta$ &
        95\% CI \\
        \midrule

        APGD-L1 &
        Uncapped &
        255 & 95.58 & 98.08 & +2.51 &
        [2.06,2.97] \\

        APGD-L1 &
        $c^\star$ &
        16 & 91.81 & 93.94 & +2.13 &
        [1.73,2.55] \\

        APGD-L1 &
        Tighter &
        8 & 86.51 & 88.55 & +2.04 &
        [1.64,2.44] \\

        \addlinespace

        FMN-L1 &
        Uncapped &
        255 & 95.96 & 96.40 & +0.44 &
        [0.19,0.72] \\

        FMN-L1 &
        $c^\star$ &
        16 & 91.89 & 92.86 & +0.97 &
        [0.69,1.28] \\

        FMN-L1 &
        Tighter &
        8 & 86.11 & 87.28 & +1.18 &
        [0.84,1.54] \\
        \bottomrule
    \end{tabular}
\end{table}

The stronger attack improves every condition. The improvements are comparable
between capped and uncapped executions rather than being concentrated at
$c=1$. At the frozen operating points, strengthened APGD-L1 and FMN-L1
retain $95.8\%$ and $96.3\%$ of their respective strengthened uncapped ASRs.
Thus, within the strengthened configurations tested here, additional
optimization changes absolute attack success while preserving the selected
high-retention behavior.

% ------------------------------------------------------------
\subsection{Frontier optimization diagnostics}
\label{app:optimizer_diagnostics}
% ------------------------------------------------------------

The exact feasible sets are nested in $c$, but the empirical finite-compute
frontiers need not be monotonic. To visualize where this occurs, we compute
\begin{equation}
\label{eq:app_delta_asr}
\Delta_{\mathrm{ASR}}(B,c)
=
\operatorname{ASR}(B,c)
-
\operatorname{ASR}(B,1)
\end{equation}
for every capped condition in the standard-model evaluation.

\begin{figure*}[t]
    \centering
    \includegraphics[width=\textwidth]
    {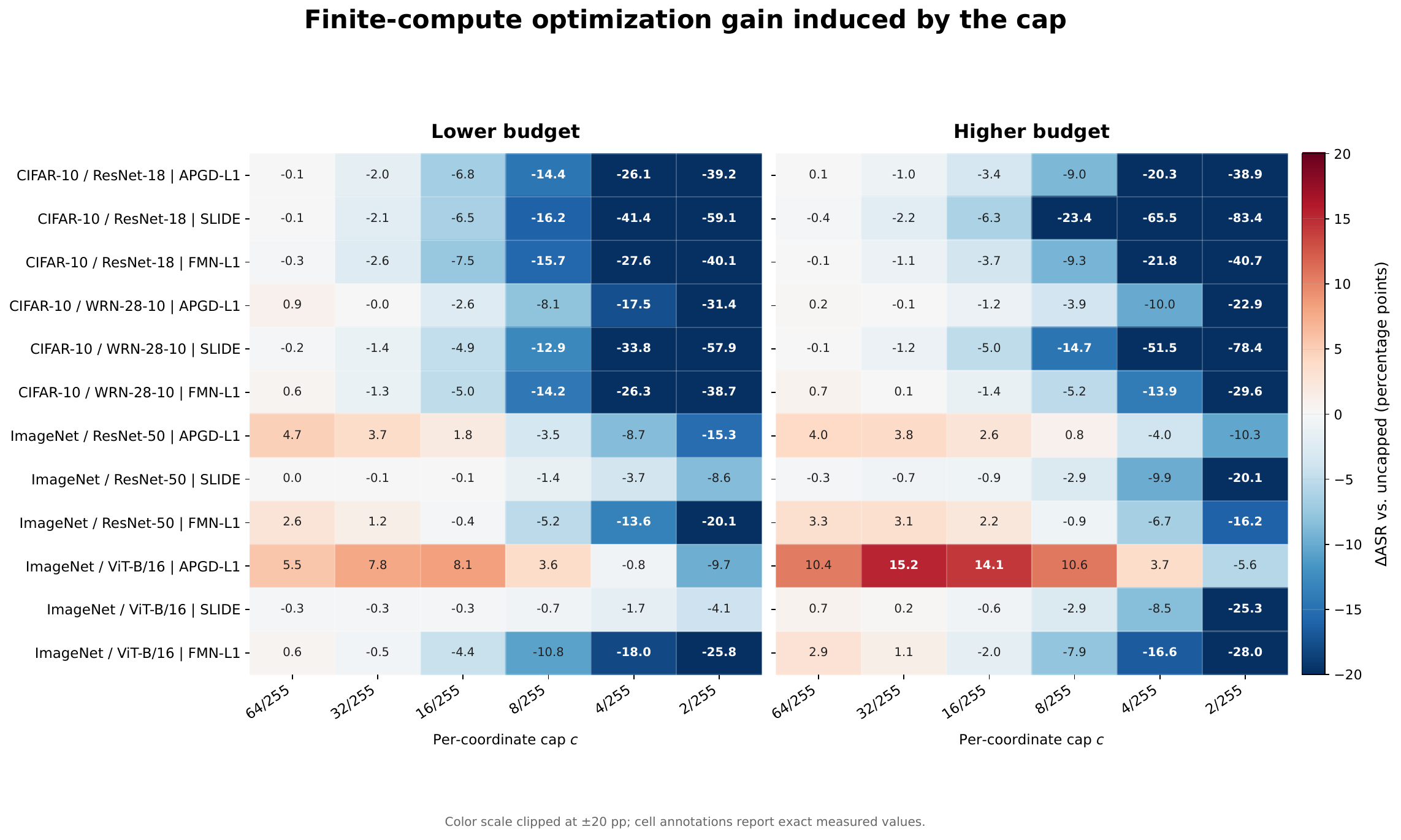}
    \caption{
    Finite-compute ASR change relative to the native uncapped reference,
    $\Delta_{\mathrm{ASR}}(B,c)$, for the CIFAR-10 and ImageNet
    standard-model frontiers. Rows correspond to
    dataset--model--attack configurations and columns to capped conditions.
    Positive entries indicate that the finite-compute capped execution
    achieves higher ASR than its directly optimized $c=1$ reference.
    No monotonic smoothing is applied. SLIDE uses the native
    $c=1$ baseline. This diagnostic localizes differences in empirical
    optimization outcomes; A06 provides a direct repeated-search audit
    of one representative discrepancy.
    }
    \label{fig:app_optimization_gain_heatmap}
\end{figure*}

\begin{figure*}[t]
    \centering
    \includegraphics[width=\textwidth]
    {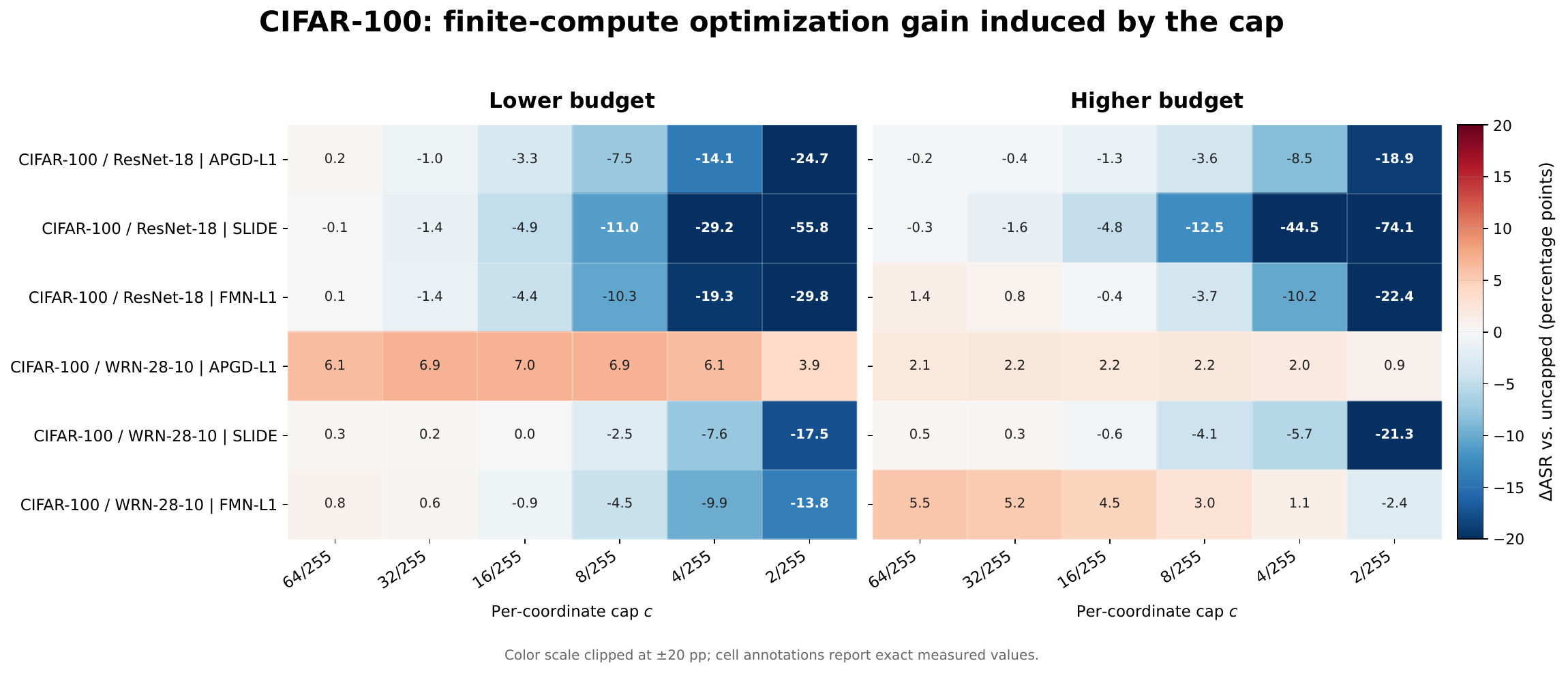}
    \caption{
    CIFAR-100 finite-compute ASR change relative to the native uncapped
    reference across both evaluated budgets. The definition and interpretation
    of $\Delta_{\mathrm{ASR}}(B,c)$ are the same as in
    Fig.~\ref{fig:app_optimization_gain_heatmap}. The figure completes the
    optimization-gain diagnostic for the standard-model evaluation rather
    than introducing an additional selection criterion.
    }
    \label{fig:app_cifar100_optimization_gain}
\end{figure*}

Figure~\ref{fig:app_optimization_gain_heatmap} reports this diagnostic for
the CIFAR-10 and ImageNet standard-model frontiers, while
Fig.~\ref{fig:app_cifar100_optimization_gain} reports the corresponding
CIFAR-100 frontiers. Positive values indicate capped executions whose
finite-compute ASR exceeds the corresponding native uncapped execution.
These cells characterize differences in empirical optimization outcomes
rather than implying that a smaller theoretical threat set is stronger.

The largest observed increase in the standard-model evaluation occurs for
APGD-L1 on ImageNet/ViT-B/16 at $B=20$, where ASR rises from $63.50\%$
at $c=1$ to $78.75\%$ at $c=32/255$, a $15.25$ percentage-point increase.
At the selected $c^\star_{0.95}=4/255$, the same configuration achieves
$67.25\%$ ASR, corresponding to $105.91\%$ retention relative to the native
uncapped execution and $85.4\%$ of the largest ASR observed on this frontier.
Because the $32/255$ perturbations also satisfy the uncapped threat-set
constraints, the distinction illustrates the implementation-relative nature
of the retention statistic. The selected operating point follows the
predefined native-relative criterion, while the complete frontier provides
the broader empirical context.

Other settings exhibit smaller positive deviations or approximately monotonic
behavior, emphasizing that empirical non-monotonicity is attack- and
model-dependent rather than a uniform property of the cap. These
measurements reflect the evaluated finite-compute optimization procedures
and do not imply a reversal of the underlying nested threat-set ordering.

The main paper also refers to this subsection for the numerical best-of-$k$
values associated with the repeated-search audit. For completeness,
Table~\ref{tab:app_restart_audit} reports those values here; the experimental
design, paired outcome analysis, and interpretation of A06 are given in
Sec.~\ref{app:restart_audit}.

\begin{table}[t]
    \centering
    \caption{
    Numerical repeated-run optimizer audit. APGD-L1 entries are cumulative
    best-of-$k$ ASR based on highest cross-entropy loss.
    FMN-L1 is an independent $500$-step reference.
    }
    \label{tab:app_restart_audit}
    \setlength{\tabcolsep}{4.0pt}
    \scriptsize
    \begin{tabular}{lrrr}
        \toprule
        Profile &
        ASR$_{c=1}$ &
        ASR$_{64/255}$ &
        $\Delta$ \\
        & (\%) & (\%) & (pp) \\
        \midrule
        APGD best-of-1 & 88.71 & 92.71 & +4.00 \\
        APGD best-of-2 & 91.14 & 93.71 & +2.57 \\
        APGD best-of-3 & 91.71 & 94.00 & +2.29 \\
        APGD best-of-4 & 91.86 & 94.14 & +2.29 \\
        APGD best-of-5 & 92.43 & 94.43 & +2.00 \\
        \midrule
        FMN, 500 steps & 90.29 & 95.00 & +4.71 \\
        \bottomrule
    \end{tabular}
\end{table}

% ------------------------------------------------------------
\subsection{Repeated-run optimizer audit: A06}
\label{app:restart_audit}
% ------------------------------------------------------------

A06 examines one finite-compute discrepancy by increasing independent
search effort on $700$ fixed clean-correct ImageNet/ResNet-50 examples at
$B=20$.

For APGD-L1, five independently seeded runs are performed under both $c=1$
and $c=64/255$. Corresponding runs use matched seeds. For every sample and
every $k\in\{1,\ldots,5\}$, the candidate with the highest cross-entropy
loss among the first $k$ runs is retained. The resulting cumulative
best-of-$k$ ASRs are reported in
Table~\ref{tab:app_restart_audit} in Sec.~\ref{app:optimizer_diagnostics}.
This loss-based selection characterizes the evaluated search procedure,
rather than aggregating success over all individual restarts.

The capped-over-uncapped APGD-L1 difference decreases from $4.00$ percentage
points after one run to $2.00$ points after five runs. Both conditions improve
with additional independent search: ASR rises from $88.71\%$ to $92.43\%$
for $c=1$ and from $92.71\%$ to $94.43\%$ for $c=64/255$.

At best-of-five, $643$ examples are attacked under both conditions, $18$
only under the capped condition, four only under the uncapped condition, and
$35$ under neither. McNemar's test gives $p\approx0.0043$.

The narrowing gap with additional independent search supports a
finite-compute interpretation of this non-monotonic frontier region. At the
same time, a residual difference remains after five runs, and the independent
$500$-step FMN-L1 reference also attains higher measured ASR at $64/255$
than at $c=1$. These observations characterize the influence of search
effort in the evaluated ResNet-50 configuration without establishing how
other frontier discrepancies would behave under stronger optimization.
The measured ordering remains dependent on the finite-compute search
procedure, consistent with the distinction between empirical attack success
and exact worst-case risk made in Sec.~\ref{sec:approach}.

The primary visualization of the repeated-search experiment is
Fig.~\ref{fig:restart_audit} in the main paper.

% ------------------------------------------------------------
\subsection{Official AutoAttack reference: A07}
\label{app:autoattack_reference}
% ------------------------------------------------------------

A07 provides an external reference for the strength of the uncapped
$c=1$ executions. We evaluate official untargeted $\ell_1$ AutoAttack on
$1{,}000$ fixed clean-correct CIFAR-10/ResNet-18 examples at $B=4$.

\begin{table}[t]
    \centering
    \caption{
    Official AutoAttack reference on the matched fixed population.
    Differences use official AutoAttack as the reference.
    }
    \label{tab:app_autoattack_reference}
    \setlength{\tabcolsep}{3.5pt}
    \scriptsize
    \begin{tabular}{lrrr}
        \toprule
        Attack &
        ASR (\%) &
        $\Delta$ (pp) &
        95\% CI \\
        \midrule
        Official $\ell_1$ AutoAttack &
        98.50 & --- & --- \\

        APGD-L1 &
        96.20 & -2.30 &
        [-3.30,-1.40] \\

        FMN-L1 &
        95.70 & -2.80 &
        [-3.90,-1.80] \\
        \bottomrule
    \end{tabular}
\end{table}

The individual frontier attacks attain slightly lower ASR than the full
official ensemble on this population. The comparison provides an external
strength reference for the uncapped APGD-L1 and FMN-L1 executions, which
remain within $2.8$ percentage points of the official ensemble under
the evaluated conditions.

% ------------------------------------------------------------
\subsection{Control coverage and interpretation}
\label{app:ablation_coverage}
% ------------------------------------------------------------

The seven controls examine distinct aspects of the main results:

\begin{itemize}
    \item \textbf{A01--A02:} whether apparent redistribution is explained by
    lower realized $\ell_1$ use;

    \item \textbf{A03:} whether the selected structural change is accompanied
    by changes in learned perceptual distance;

    \item \textbf{A04:} whether post-hoc clipping can reproduce direct capped
    optimization;

    \item \textbf{A05:} whether the selected operating points retain their
    effectiveness under stronger attack configurations;

    \item \textbf{A06:} how a representative capped-over-uncapped ASR
    discrepancy responds to additional independent search;

    \item \textbf{A07:} how native uncapped frontier attacks compare with a
    standard official $\ell_1$ attack suite.
\end{itemize}

The controls address complementary aspects of the evaluation rather than
repeated versions of a single robustness check. A01--A02 assess whether
structural differences persist at matched realized perturbation magnitude;
A03 characterizes associated changes in learned perceptual distance;
A04 compares direct constrained optimization with post-hoc processing; and
A05--A07 examine optimization strength and uncapped reference quality
within their respective evaluated settings.

Together, these experiments support the interpretation of the measured
concentration frontiers while clarifying the roles of perturbation magnitude,
constraint enforcement, and finite-compute optimization. Their conclusions
are specific to the evaluated models, attacks, and control configurations;
they do not require exact inner maximization or imply that the empirical
frontiers represent optimal adversarial risk. In particular, observed
non-monotonicities characterize finite-compute attack behavior rather than
the ordering of the underlying threat sets.

\section{Additional Frontier Analyses}
\label{app:additional_frontier}

This appendix collects complementary analyses of the completed
perturbation--concentration frontiers. None of the results in this section is
used to select or revise the primary $c^\star_{0.95}$ operating points.
Instead, they examine the sensitivity of the discrete operating-point summary
to the retention convention and global budget, complete the visual reporting
of the CIFAR-100 standard-model frontiers, and characterize descriptive
associations between perturbation structure and learned perceptual distance.

% ------------------------------------------------------------
\subsection{Retention-threshold sensitivity}
\label{app:threshold_sensitivity}
% ------------------------------------------------------------

The primary analysis defines $c^\star$ using a retention threshold of
$\rho=0.95$. To test whether the resulting concentration scale is specific
to this convention, we recompute the operating-point summary on the already
completed frontiers at
\[
\rho\in\{0.90,0.95,0.99\}.
\]
No attack is rerun for this analysis, and the original
$c^\star_{0.95}$ values remain frozen for all downstream controls. The
standard-model summaries below use the native $c=1$ SLIDE reference
executions.

Table~\ref{tab:app_threshold_sensitivity} summarizes the resulting
distributions at the reference budgets. For the $18$ standard-model
attack--model settings, the median selected cap is $8/255$ under both
$90\%$ and $95\%$ retention. Requiring $99\%$ retention moves the median
to $32/255$. The largest selected standard-model cap similarly moves from
$16/255$ at $90\%$ retention to $32/255$ at $95\%$ and $64/255$ at
$99\%$.

\begin{table}[t]
    \centering
    \caption{
    Sensitivity of the selected operating-point distribution to the retention
    threshold at the reference budgets. Entries in the distribution column
    give counts of settings selecting each value of $255c^\star_\rho$.
    Standard results cover the $18$ standard attack--model settings; robust
    results cover the eight APGD-L1/SLIDE reference-budget settings.
    The primary paper uses $\rho=0.95$.
    }
    \label{tab:app_threshold_sensitivity}
    \scriptsize
    \setlength{\tabcolsep}{3.5pt}
    \begin{tabular}{lllc}
        \toprule
        Regime & $\rho$ &
        Distribution of $255c^\star_\rho$ &
        Median \\
        \midrule

        Standard
        & 0.90
        & $2{:}3,\ 4{:}4,\ 8{:}7,\ 16{:}4$
        & 8 \\

        Standard
        & 0.95
        & $2{:}2,\ 4{:}2,\ 8{:}7,\ 16{:}4,\ 32{:}3$
        & 8 \\

        Standard
        & 0.99
        & $2{:}1,\ 4{:}2,\ 8{:}1,\ 16{:}4,\ 32{:}5,\ 64{:}5$
        & 32 \\

        \addlinespace

        Robust
        & 0.90
        & $4{:}1,\ 16{:}1,\ 32{:}3,\ 64{:}1,\ 255{:}2$
        & 32 \\

        Robust
        & 0.95
        & $4{:}1,\ 32{:}1,\ 64{:}4,\ 255{:}2$
        & 64 \\

        Robust
        & 0.99
        & $4{:}1,\ 64{:}2,\ 255{:}5$
        & 255 \\

        \bottomrule
    \end{tabular}
\end{table}

Relative to the primary $95\%$ criterion, lowering the standard-model
threshold to $90\%$ selects a tighter cap in $10$ of $18$ settings and the
same cap in the remaining eight. Increasing the threshold to $99\%$ selects
a looser cap in $16$ settings and the same cap in two. This behavior is
expected: requiring a larger fraction of the measured uncapped ASR makes the
retention requirement harder to satisfy on the fixed cap grid.

The standard-model conclusion is therefore not specific to the distinction
between $90\%$ and $95\%$ retention: both criteria select a median cap of
$8/255$. The $99\%$ criterion is substantially more demanding and shifts
the operating points toward weaker coordinate restrictions, but even there
the standard-model median remains below the conventional $c=1$ reference.

The robust-model frontiers show markedly greater sensitivity. Their median
selected cap moves from $32/255$ at $\rho=0.90$ to $64/255$ at
$\rho=0.95$ and to the uncapped value at $\rho=0.99$. This reinforces the
main-paper observation that adversarially trained classifiers exhibit
heterogeneous concentration sensitivity. The robust results are not pooled
with the standard-model results and are reported here only as a
reference-budget sensitivity summary.

% ------------------------------------------------------------
\subsection{Budget sensitivity of the operating point}
\label{app:budget_sensitivity}
% ------------------------------------------------------------

Because $c^\star_{0.95}$ is selected independently at each global budget, the
two-budget standard-model frontiers also allow us to ask whether the amount of
permitted concentration associated with $95\%$ retention changes with the
available global $\ell_1$ magnitude.

Table~\ref{tab:app_budget_sensitivity} reports the complete standard-model
comparison. Each entry gives
\[
255c^\star_{0.95}(B_{\mathrm{low}})
\rightarrow
255c^\star_{0.95}(B_{\mathrm{high}}).
\]
The budget pair is $2\rightarrow4$ for CIFAR-10 and CIFAR-100 and
$10\rightarrow20$ for ImageNet. Smaller cap values represent stronger
restrictions on individual coordinate magnitude.

\begin{table}[t]
    \centering
    \caption{
    Budget dependence of the $95\%$-retention operating point for all
    standard models. Smaller values correspond to tighter coordinate caps.
    Across the $18$ model--attack comparisons, the higher budget selects a
    tighter cap in $10$ cases and the same cap in eight; no comparison
    becomes looser.
    }
    \label{tab:app_budget_sensitivity}
    \scriptsize
    \setlength{\tabcolsep}{2.8pt}
    \begin{tabular}{llccc}
        \toprule
        Dataset & Model &
        APGD-L1 & SLIDE & FMN-L1 \\
        \midrule

        CIFAR-10
        & ResNet-18
        & $32\!\rightarrow\!16$
        & $32\!\rightarrow\!32$
        & $32\!\rightarrow\!16$ \\

        CIFAR-10
        & WRN-28-10
        & $16\!\rightarrow\!8$
        & $32\!\rightarrow\!32$
        & $32\!\rightarrow\!16$ \\

        CIFAR-100
        & ResNet-18
        & $16\!\rightarrow\!8$
        & $32\!\rightarrow\!32$
        & $32\!\rightarrow\!8$ \\

        CIFAR-100
        & WRN-28-10
        & $2\!\rightarrow\!2$
        & $8\!\rightarrow\!8$
        & $16\!\rightarrow\!2$ \\

        ImageNet
        & ResNet-50
        & $16\!\rightarrow\!4$
        & $8\!\rightarrow\!8$
        & $16\!\rightarrow\!8$ \\

        ImageNet
        & ViT-B/16
        & $4\!\rightarrow\!4$
        & $8\!\rightarrow\!8$
        & $32\!\rightarrow\!16$ \\

        \bottomrule
    \end{tabular}
\end{table}

Across the $18$ matched model--attack comparisons, increasing the global
budget produces a tighter selected cap in $10$ settings and leaves it
unchanged in eight. No setting requires a looser cap at the higher budget.
For SLIDE, the selected operating point remains unchanged across budgets
for all six standard models, although its absolute value varies across
models.

This pattern should not be interpreted as a universal monotonic law linking
$B$ and $c^\star$. The operating point is a discrete empirical summary that
depends on the attack, classifier, cap grid, and finite-compute optimization.
Within the evaluated standard-model settings, however, the higher global
budget is consistently associated with an equal or tighter $95\%$-retention
cap. One plausible interpretation is that additional global $\ell_1$ mass
provides greater scope for redistribution when individual coordinate
magnitudes are restricted; the present results establish the empirical
pattern, not that mechanism.

% ------------------------------------------------------------
\subsection{Complete CIFAR-100 standard-model frontiers}
\label{app:cifar100_frontiers}
% ------------------------------------------------------------

\begin{figure*}[t]
    \centering
    \includegraphics[width=\textwidth]
    {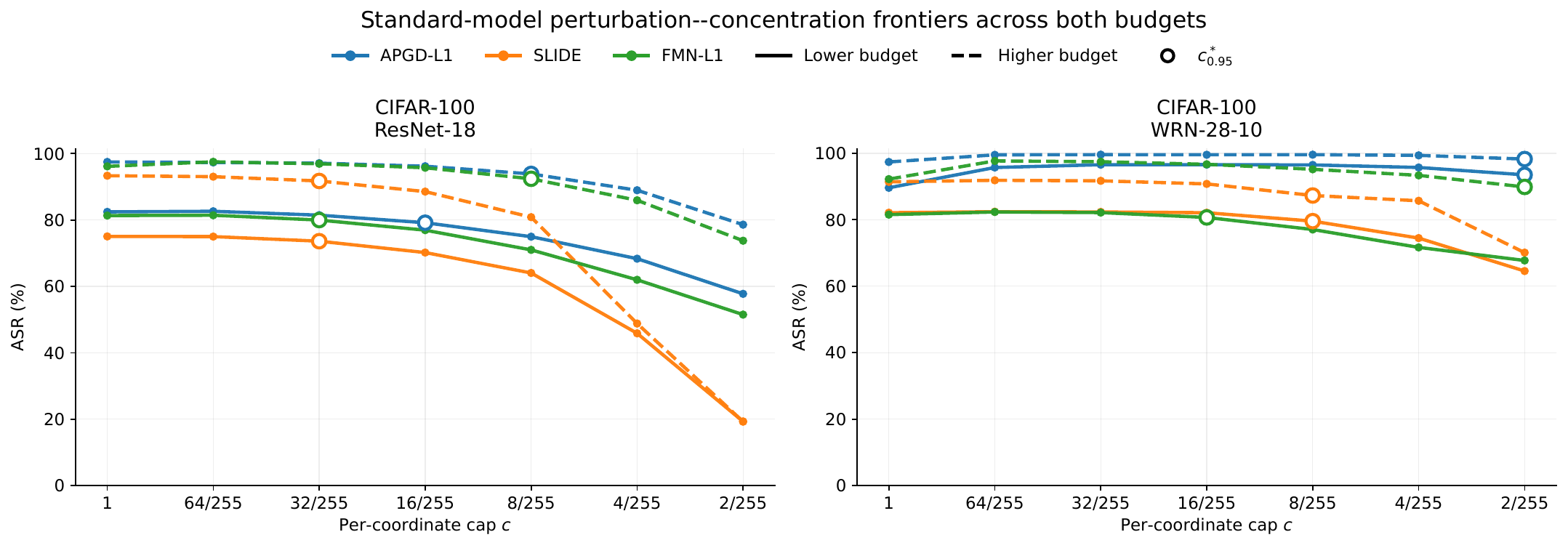}
    \caption{
    Complete CIFAR-100 perturbation--concentration frontiers across both
    evaluated global budgets for ResNet-18 and WRN-28-10. Curves show
    APGD-L1, SLIDE, and FMN-L1 as the per-coordinate cap is tightened over
    the same predefined grid used in the main evaluation. Solid and dashed
    curves distinguish the two global budgets, and markers denote the
    independently selected $c^\star_{0.95}$ operating points. No monotonic
    smoothing is applied. The SLIDE curves use the native $c=1$
    reference executions.
    }
    \label{fig:app_cifar100_frontiers}
\end{figure*}

The main paper reports the primary CIFAR-100 operating points together with
the other standard-model settings but visualizes the complete two-budget
frontiers for CIFAR-10 and ImageNet. Figure~\ref{fig:app_cifar100_frontiers}
provides the corresponding complete CIFAR-100 visualization using the
native SLIDE $c=1$ reference executions.

The CIFAR-100 results reproduce the main qualitative behavior on a second
label space at the same image resolution as CIFAR-10. At the reference
budget $B=4$, all six model--attack settings satisfy the $95\%$ retention
criterion at caps no larger than $32/255$, with several substantially tighter
operating points. At the same time, the frontier shapes remain
attack- and model-dependent, including finite-compute non-monotonic regions.
The complete curves are therefore more informative than treating the selected
$c^\star$ values as universal concentration thresholds.

% ------------------------------------------------------------
\subsection{Associations between structure and perceptual distance}
\label{app:metric_correlations}
% ------------------------------------------------------------

The perceptual analysis in Sec.~\ref{app:perceptual_results} directly
compares prespecified uncapped, selected-cap, and tighter-cap conditions.
As a complementary descriptive analysis, we also measure the pooled
association between perturbation-structure quantities and learned perceptual
distance over the successful ImageNet/ResNet-50 perceptual-evaluation rows.

Table~\ref{tab:app_metric_correlations} reports Spearman rank correlations
with $95\%$ intervals obtained by resampling complete sample-ID clusters.
The analysis contains $5{,}835$ successful condition rows from $720$ unique
sample clusters.

\begin{table}[t]
    \centering
    \caption{
    Descriptive Spearman correlations between perturbation-structure metrics
    and LPIPS/DISTS over successful ImageNet/ResNet-50 perceptual-evaluation
    rows. Brackets give $95\%$ sample-cluster bootstrap intervals. The
    estimates pool attacks and cap conditions and are not interpreted
    causally.
    }
    \label{tab:app_metric_correlations}
    \scriptsize
    \setlength{\tabcolsep}{4.0pt}
    \begin{tabular}{lcc}
        \toprule
        Structural metric &
        LPIPS $\rho_s$ [95\% CI] &
        DISTS $\rho_s$ [95\% CI] \\
        \midrule

        $\ell_\infty$
        & $0.344\ [0.326,0.365]$
        & $0.418\ [0.400,0.439]$ \\

        Top-$1\%$ mass
        & $-0.301\ [-0.327,-0.274]$
        & $-0.218\ [-0.249,-0.187]$ \\

        Top-$5\%$ mass
        & $-0.051\ [-0.076,-0.024]$
        & $0.012\ [-0.014,0.037]$ \\

        $S_{\mathrm{eff}}$
        & $0.130\ [0.103,0.155]$
        & $0.076\ [0.045,0.106]$ \\

        \bottomrule
    \end{tabular}
\end{table}

Peak magnitude has the clearest positive pooled association with both learned
perceptual-distance measures, whereas the top-$5\%$ mass fraction is only
weakly associated with LPIPS and is statistically compatible with no
association for DISTS. The signs of the remaining associations are not
sufficient to define a single perceptual concentration score.

These pooled estimates combine different attacks and cap conditions and
therefore reflect both within-condition and between-condition variation.
They should not be interpreted as evidence that changing any one structural
metric causes a corresponding perceptual change. In particular, the
correlation analysis does not replace the paired common-success comparisons
in Sec.~\ref{app:perceptual_results}. Rather, it supports the decision to
report peak magnitude, top-mass concentration, effective support, LPIPS, and
DISTS as distinct quantities instead of treating any one of them as a proxy
for the others.

\end{document}